\documentclass[english]{IEEEtran}
\usepackage[T1]{fontenc}
\usepackage[latin9]{inputenc}
\usepackage{array}
\usepackage{float}
\usepackage{multirow}
\usepackage{graphicx}
\usepackage{amssymb}
\usepackage{graphicx}
\usepackage{cite}
\usepackage{amsmath}
\usepackage{array}
\usepackage{multirow}
\usepackage{babel}
\usepackage{amsmath}
\usepackage{url}



\begin{document}

\title{Driver distraction detection and recognition using RGB-D sensor}

\author{C\'eline Craye \and and Fakhri Karray\\
Electrical and Computer Engineering\\
University of Waterloo\\
ccraye@uwaterloo.ca}
\maketitle

\begin{abstract}

Driver inattention assessment has become a very active field in intelligent transportation systems. Based on active sensor Kinect and computer vision tools, we have built an efficient module for detecting driver distraction and recognizing the type of distraction. Based on color and depth map data from the Kinect, our system is composed of four sub-modules. We call them eye behavior (detecting gaze and blinking), arm position (is the right arm up, down, right of forward), head orientation, and facial expressions. Each module produces relevant information for assessing driver inattention. They are merged together later on using two different classification strategies: AdaBoost classifier and Hidden Markov Model. Evaluation is done using a driving simulator and 8 drivers of different gender, age and nationality for a total of more than 8 hours of recording. Qualitative and quantitative results show strong and accurate detection and recognition capacity (85\% accuracy for the type of distraction and 90\% for distraction detection). Moreover, each module is obtained independently and could be used for other types of inference, such as fatigue detection, and could be implemented for real cars systems.

\end{abstract}

\section{Introduction}

Alarming statistics about distracted driving can be found on the official US government website about distracted driving \cite{distraction.gov}. In 2010, 18\% of injury crashes were distraction-related. 3331 people were killed in 2011 in a crash involving a distracted driver, and distraction is responsible for 11\% of fatal crashes of drivers under the age of twenty. These statistics are even more worrying as the number of possible distractions within a car keeps increasing. The large number of displays and new infotainment devices in cars has made the problem more critical. \\

The National Highway Transportation Safety Administration (NHTSA) has defined distracted driving as ``an activity that could divert a person's attention away from the primary task of driving''. It is commonly classified into 3 categories, namely
\begin{itemize}
\item Manual distraction: The driver takes his hands off the wheel. This includes for example text messaging, eating, using a navigation system, or adjusting radio.
\item Visual distraction: The driver takes his eyes off the road, for example reading or watching a video.
\item Cognitive distraction: The driver's mind is not fully focused on driving. This can happen when the driver is talking to other passengers, texting, or simply thinking.
\end{itemize}

The influence of distraction on drivers performance has been widely studied \cite{stutts2003driver}, \cite{strayer2004profiles}, \cite{blueprint}, \cite{Trafficsafetyfacts}, \cite{olson2009driver}, and interesting facts have come to light: cell phones use represent 18\% of distracted fatal driver accident in North America. Indeed, cell phone conversations induce a high level of cognitive distraction, thus reducing the brain activity related to driving by 37\% (which might be worse than ingesting alcohol). Handsfree cell phones have not been found particularly safer that hand-held use. More importantly, text messaging requires visual, manual, and cognitive attention at the same time, making it the most dangerous distraction. It was found that text messaging takes the driver's eyes off the road for 4.6 seconds, which is sufficient to drive the length of a football field completely blind. The crash risk when text messaging is twenty-three times worse than driving with no distraction. \\

All these facts suggest that drivers should be aware of the risk, but it is also the car manufacturer's responsibility to offer intelligent assistance tools to avoid driver distraction, and to limit crash risks. This issue is still an open problem, as the variety of actions, the differences between drivers and outdoor conditions make this task extremely challenging.\\

Our approach aims at determining first if a driver is distracted or not, and in the case he is, the system should be able to recognize the type of distraction. Based on computer vision techniques, we propose four different modules for features extraction, focusing on arm position, face orientation, facial expression and eye behavior. We propose two strategies to combine the output information from each module: an AdaBoost classifier with temporal smoothing, and a Hidden Markov Model-based classifier.\\

This paper is organized as follows: Section \ref{sec:lit_rev} is an overview of existing methods and commercial products. Section \ref{sec:modules} presents in detail each module for assessing driver distraction. Section \ref{sec:fusion} describes our fusion strategies for distraction recognition, section \ref{sec:experimental_results} shows our main experimental results and demonstrates the efficiency of our approach. Last section \ref{sec:conclusion} is a concluding section, discussing our results and highlights our future work.

\section{Literature review}
\label{sec:lit_rev}

In this section, we review the existing studies that have been carried out in the field of driver inattention, and we discuss their strengths and shortcomings. We only present here approaches that were useful for our study, or having a significant impact on the community. For a more extensive survey, some literature reviews can be found in: \cite{williamson2005review}, \cite{young2007driver}, \cite{dong2011driver}.\\

First of all, an important distinction should be made between driver inattention, driver fatigue and driver distraction. They all alter driver performance, but they are not caused by the same factors and can have various effects on the driver behavior. Fatigue is related to drowsiness and is affecting the driver because of physical, physiological or psychological reasons. Distraction was defined earlier and is related to an object, a person, an idea or an event that diverts the driver. Fatigue and distraction are both considered as driver inattention. A precise definition and relationship analysis between those terms has been attempted by Regan et al. \cite{regan2011driver}. Our work deals exclusively with driver distraction, and most of existing methods apply either for driver fatigue or driver inattention in general. As many techniques, features and fusion strategies are similar for fatigue, inattention and distraction, our review considers both fatigue and distraction detection methods.\\

Three main categories of system have been used for determining driver inattention, and a few studies have used a hybrid approach to combine them: 

\subsubsection{Physiological sensors}
This approach detects physiological features such as brain activity, heart rate or hand moist \cite{shiwu2011active}, \cite{lal2002driver}. In particular, electroencephalograph (EEG) has been found to be a valid, objective and accurate measure of driver inattention using $\delta$, $\theta$, $\alpha$, $\beta$ and $\sigma$ brain wave activity. However, physiological sensors are intrusive and cannot be used inside a car for commercial applications. They can be used as ground truth for studies, but they do not represent a realistic solution for driver inattention monitoring.

\subsubsection{Driver performance}
This approach uses external information and indicators of driver performance to infer the level of inattention. This includes, for example lateral position, steering wheel movements or pedal activity. For example, Raynney et al. \cite{ranney2008driver} suggested that distraction involved an sensitive lack of vehicle control, such as drifting from a side of the road, or unexpected speed changes. In 2007, Volvo introduced the Driver Alert Control system \cite{Volvo}, monitoring the road continuously with a camera and alerting the driver in case of dangerous behavior. These methods are correlated with driver inattention \cite{pilutti1997identification}, but they are also affected by external factors such as driver experience, road type, weather and outdoor light. Moreover, the measures rely on long term statistics and the system is unable to predict immediate dangers such as micro-sleep events.

\subsubsection{Computer vision}

The third approach, probably the most popular, relies on visual driver features. When inattentive, the driver's face and body show characteristic behaviors. Placing a camera in front of the driver and analysing his face expressions and movements makes a lot of sense and is considered as an efficient way for assessing driver inattention. In particular, the so-called eyes-off-road glance duration and head-off-road glance time are recognized as valid measures for visual distraction and can be assessed using an embedded camera \cite{angell2006driver}. For driver fatigue, PERCLOS (percentage of eye closure) \cite{dinges1998perclos} is considered as the best correlated physiological feature. Other behaviors such as yawning \cite{ji2006probabilistic}, \cite{Ji2004},
 \cite{bergasa2008analysing} or nodding \cite{Bergasa2006}, \cite{Senaratne2007} are also popular features, widely used in the field. Existing systems usually rely on a hardware setup that can be a simple color camera \cite{smith2003determining}, \cite{DOrazio2007}, \cite{Senaratne2007}, an infrared camera able to alternate between bright and dark pupil effect (useful for eye detection and robust to illumination variations) \cite{Bergasa2006}, \cite{ji2006probabilistic}, \cite{Ji2004}, \cite{craye2013multi} or a set of cameras to improve face orientation estimation \cite{bergasa2008analysing}. Recently, the Microsoft Kinect sensor has received particular attention \cite{limulti2012} as it provides both color camera, depth map and comes with a powerful face tracker. Image processing and computer vision tools are then used for extracting indicators of inattention. The key components are face, eyes and mouth detection. This is the starting point for feature extraction such as yawning, nodding, face motion, gaze estimation, or blink detection. Statistics such as PERCLOS, gaze distribution, yawning or nodding frequencies can be computed based on the features to infer driver inattention.  Last, a fusion module is designed in order to merge the information and infer the level of inattention. Most popular fusion techniques are fuzzy logic \cite{Bergasa2006}, \cite{damousis2008fuzzy}, \cite{Senaratne2007}, Bayesian networks \cite{Ji2004} and dynamic Bayesian networks \cite{ji2006probabilistic}, neural networks \cite{limulti2012} or simple decision rules \cite{smith2003determining}, \cite{bergasa2008analysing}.
Recently, commercial products, such as Eye Alert \cite{eyeAlert} have been conceived to detect driver inattention using computer vision and emit a warning in case of dangerous situation.

\subsubsection{Hybrid systems}
Last, a few approaches use a combination of the three aforementioned techniques. For example, Daza et al. \cite{daza2011drowsiness} used both driver (PERCLOS, nodding, etc.) and driving (lane variation, steering wheel movements, etc.) data to assess fatigue. Fletcher et al. \cite{fletcher2005correlating} have successfully merged driver gaze and road information to detect if a driver was missing any road sign. Last Li et al. \cite{limulti2012} have used both computer vision, steering wheel signal and pulse oxymeter to infer driver fatigue.

\subsubsection{Proposed approach and contributions}
Among state of the art techniques, most of them focus on driver fatigue detection, sometimes extended to driver inattention. To our knowledge, no serious study has been done on distraction only, trying to detect the type of action the driver is accomplishing. Determining the type of driver distraction provides higher level information than just the level of distraction. It could be used for number of applications related to intelligent transportation systems. For inter vehicles communication, providing the action a driver is doing can be more explicit and useful than statistics on how distracted that driver is. Even for smart cars, detecting the type of distraction enables statistics computation on the driver's behaviour that could further help the vehicle in keeping the driver safe.\\

Our system is based on a Kinect sensor. Originally conceived for entertainment and video game applications, it has quickly become a good tool for the computer vision community. Indeed, not only it was the first low-cost depth sensor for general public, but it also came with a very polished SDK, giving developers a large range of possibilities. For example, the SDK provides a quite efficient skeletal tracking algorithm and tools for gesture recognition. In our case, the RGBD (RGB-depth) data is very helpful for driver segmentation as well as face detection and tracking. To the best of our knowledge, only Li et al. \cite{limulti2012} have published work making use of the Kinect for car safety applications.\\

As most of existing systems rely solely on driver's face behavior, we also use driver's gesture to help us is our inference task. Unlike traditional approaches, our sensor is placed in such a way that driver's upper body is visible. Thus, we can extract driver arms position and motion. This feature will be of major help for determining driver distraction.

\section{Feature extraction}
\label{sec:modules}

This section explains how to extract features from each body and face components. We divide our task into four independent modules, namely (1) arm position estimation, (2) face orientation, (3) facial features - called animation units (AUs) - such as mouth shape and eyebrow raising, and (4) gaze estimation and eye closure. Each module uses either depth data, color data, or both. They are fused later on via different fusion schemes to determine the type of distraction. The next sections describe in detail the realization of each module.

\subsection{Arm position}

\begin{figure*}
	  \centering
		\includegraphics[scale=0.6]{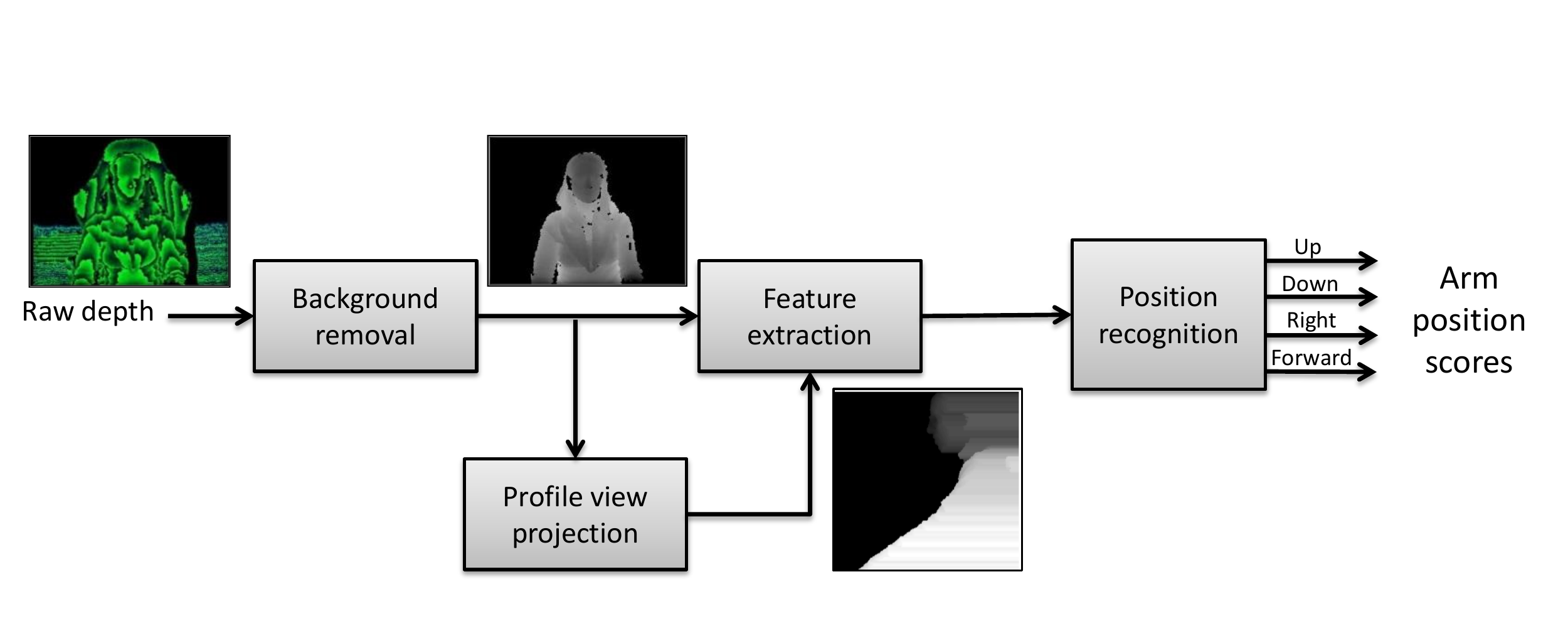}
    \caption{Main blocks of the arm position module. The input is the raw depth data. After processing, we extract features and use an AdaBoost classifier to output four scores representing the arm position.}
    \label{im:arm_position}
\end{figure*}

Much effort have been deployed by Microsoft on providing an efficient skeleton tracking to help developers in gesture recognition, both for the entire and the half body (in seated mode) \cite{held20123d}. However, the algorithm has been tested on our recorded sequences, and has shown poor results, even in seated mode. The main reasons for this is the shape of the seat of our simulator, which is hard to differenciate from the body, and the relatively small motion of the driver during a driving session. Other methods and source code are available for arm tracking or pose estimation, such as \cite{PuppetParade}, \cite{zhu2007constrained}, or \cite{shotton2013real}, but none of these approach was satisfying for us, either because they required a large training set, or because we obtained poor performance on our sequences.

For this reason, we decided to build our own arm detection system within the framework of machine learning for arm position recognition. Our features are based on the depth map data and represent the orientation of the rims of the right arm of the driver. Figure \ref{im:arm_position} shows the main steps of our method.\\

\subsubsection{Background removal}
First of all, a background/foreground segmentation is applied to isolate the driver from the driving seat and background. Kinect depth data already contains 3 bits of player data, representing the location of each player, thus providing rough segmentation. As a more accurate segmentation is required in our case, we have used a background subtraction-based approach. First, short video sequences of the seat without the driver is recorded, and an average depth map of the seat only is obtained. Then, this depth map is subtracted from the depth map of driving sequences, and image thresholding is applied to remove the seat and the background. We then combine Kinect segmentation with our technique, and we clean the irrelevent blobs by keeping the biggest one only.\\

\subsubsection{Feature extraction}
Based on the segmented depth map, we now need to find features for discriminating arm position. Most of the time, the Kinect records the driver with a frontal view, and the right arm is therefore on the right side of the body. Based on this assumption, we extract features based on foreground contours. We first apply to the binary foreground image the marching squares algorithm - the 2D version of the marching cubes \cite{lorensen1987marching} - which provides us an ordered list of contour pixels (in this list, each pixel is preceded and followed by neighboring pixels in the image). Then, we only consider the pixels of the right side of the body by removing from the list the pixels on the left hand side. Those pixels represent the rim of the right head, shoulder, arm and body of the driver. \\

We then cut this ``half'' contour list into twenty successive segments, composed of the same number of pixels. Each pixel of the list is associated with a 3D point (given image coordinates and depth value), such that each segment of the list corresponds to a 3D point cloud. We then extract for each segment the main axis - a 3D vector - of the associated point cloud by principal component analysis. Putting together all the main axis, we obtain a $20 \times 3$ feature vector.\\

\begin{figure*}
\centering
\begin{tabular}{cccc}
	 &\includegraphics[scale=0.25]{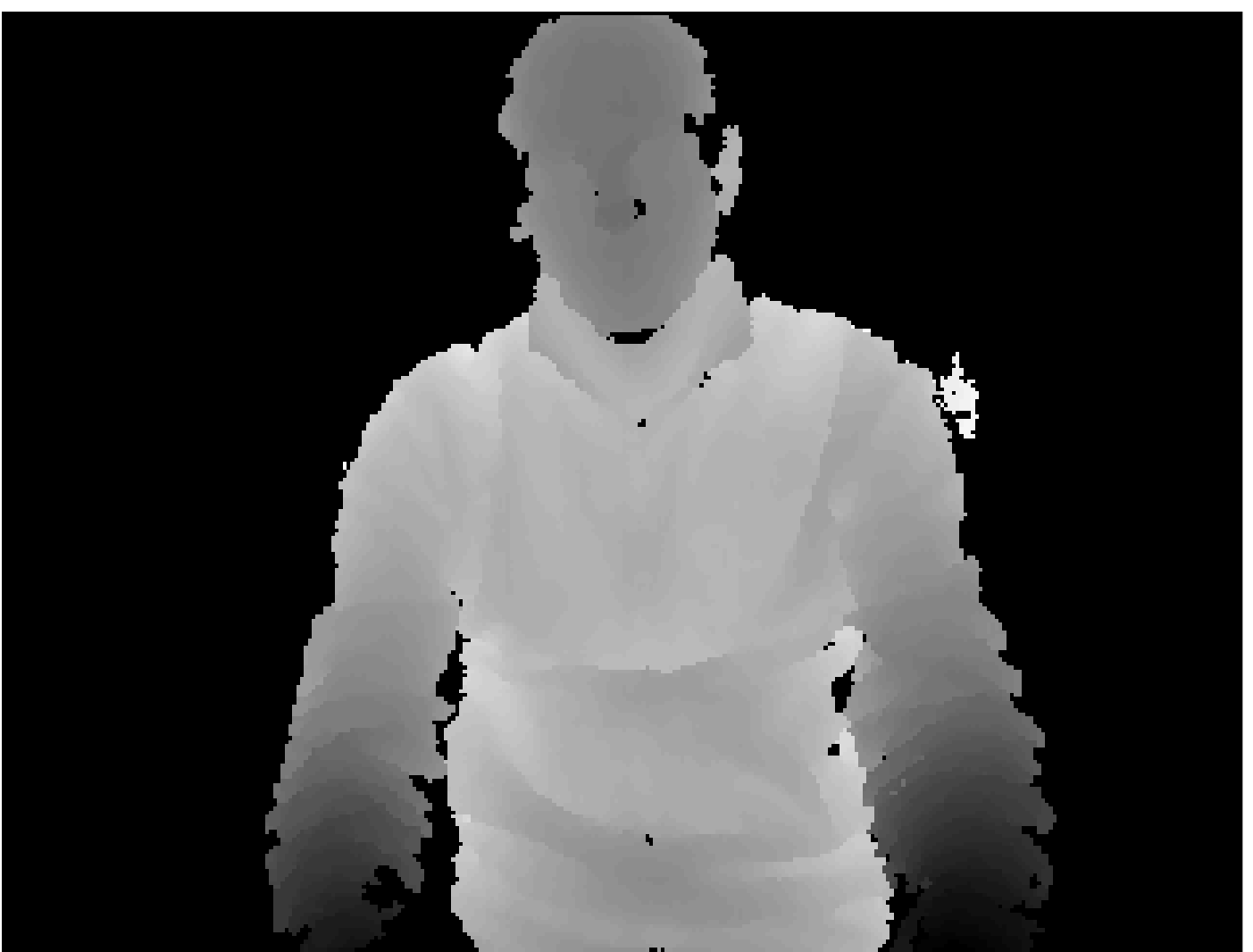} &
   \includegraphics[scale=0.25]{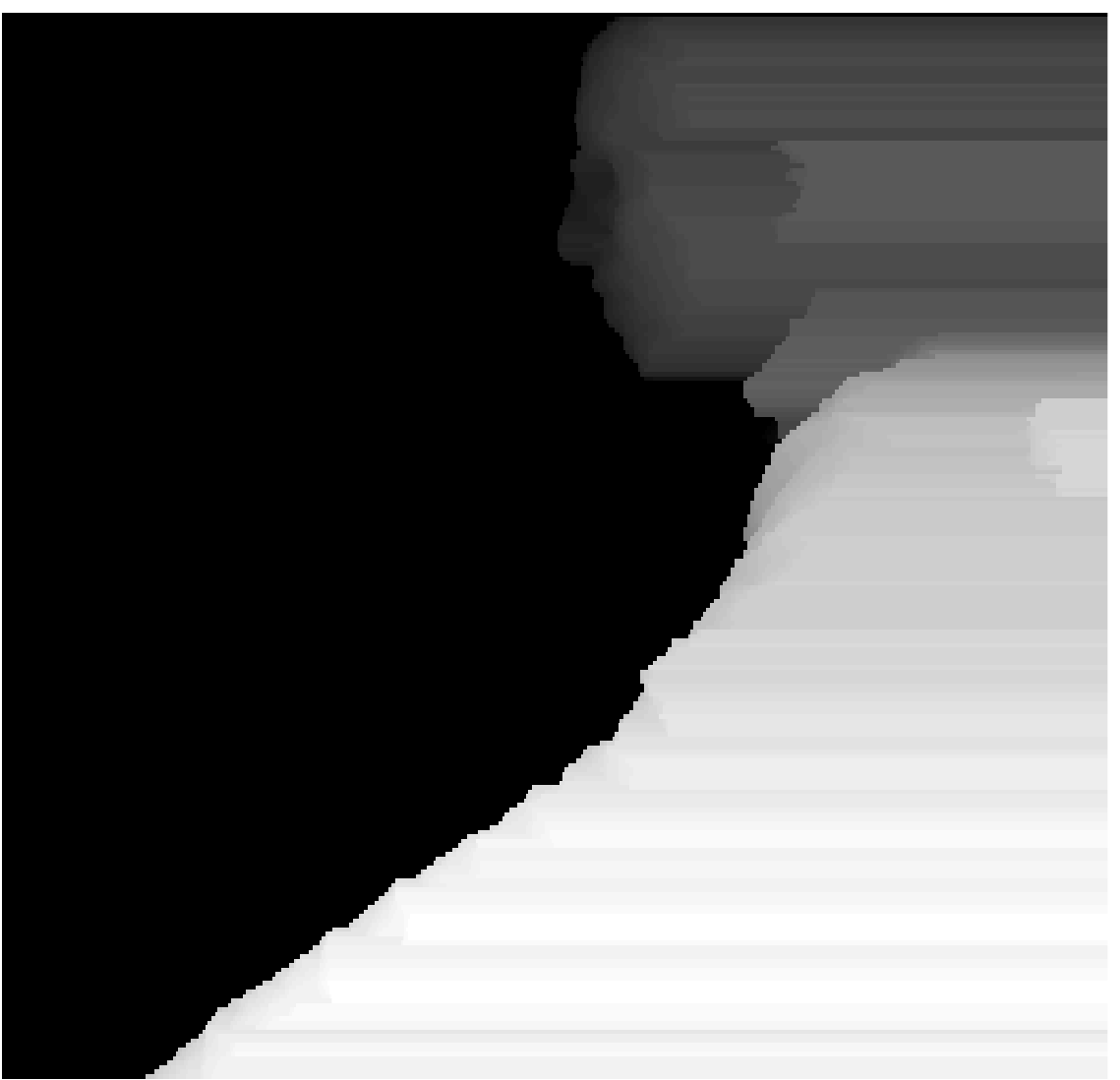}& \\
    & Frontal view & Profile view & \\
   \includegraphics[scale=0.25]{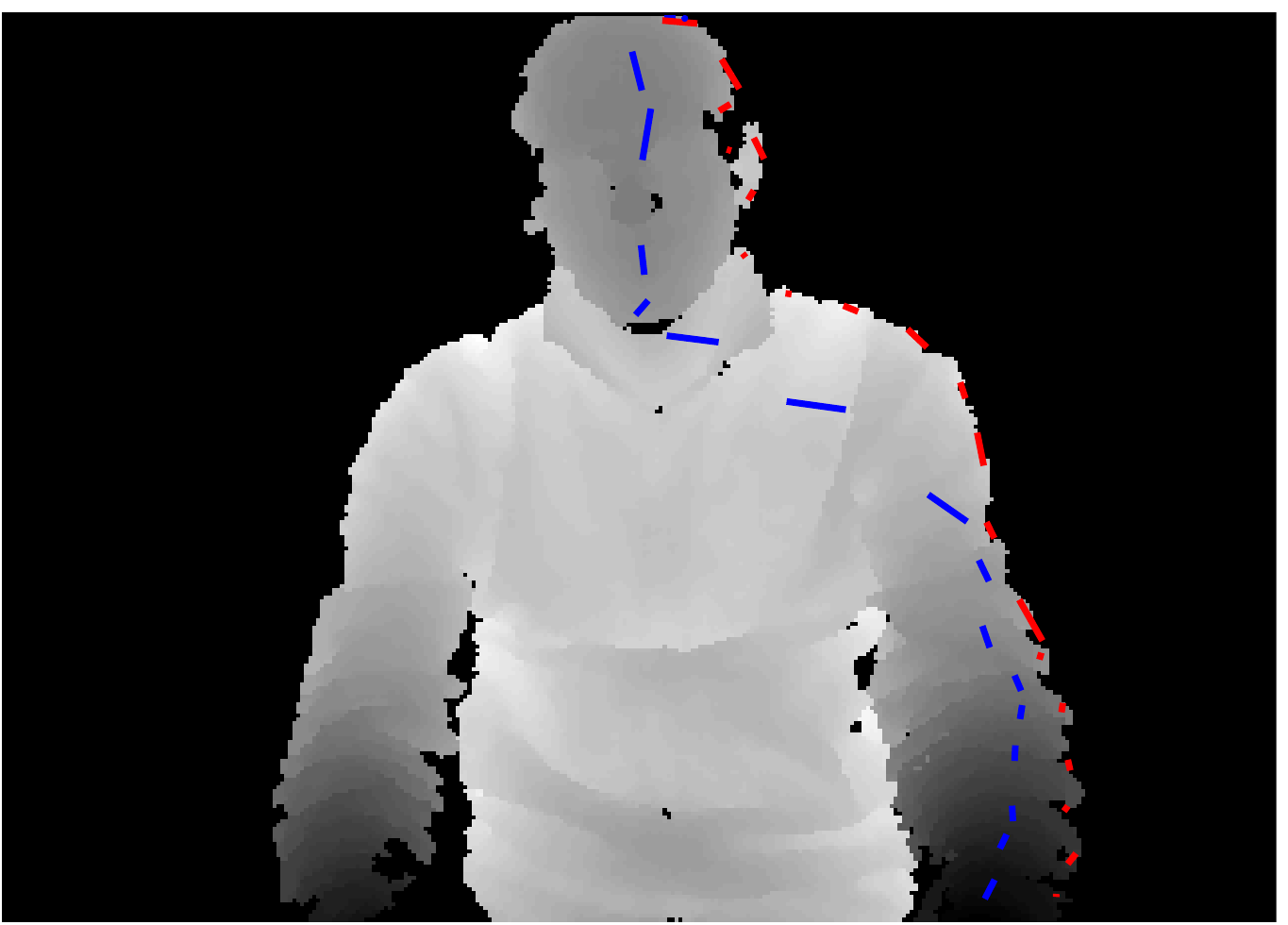} &
   \includegraphics[scale=0.25]{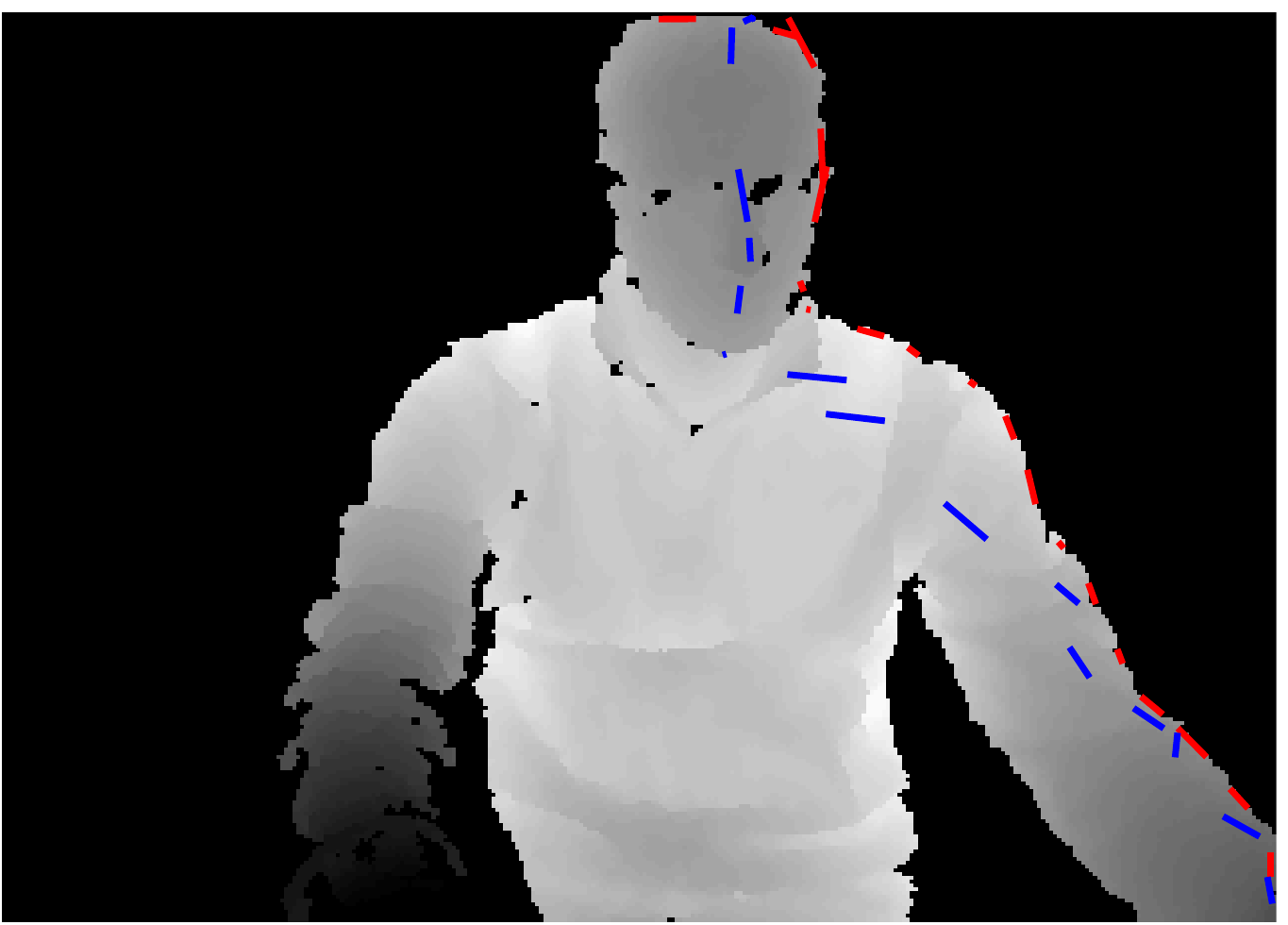} &
   \includegraphics[scale=0.25]{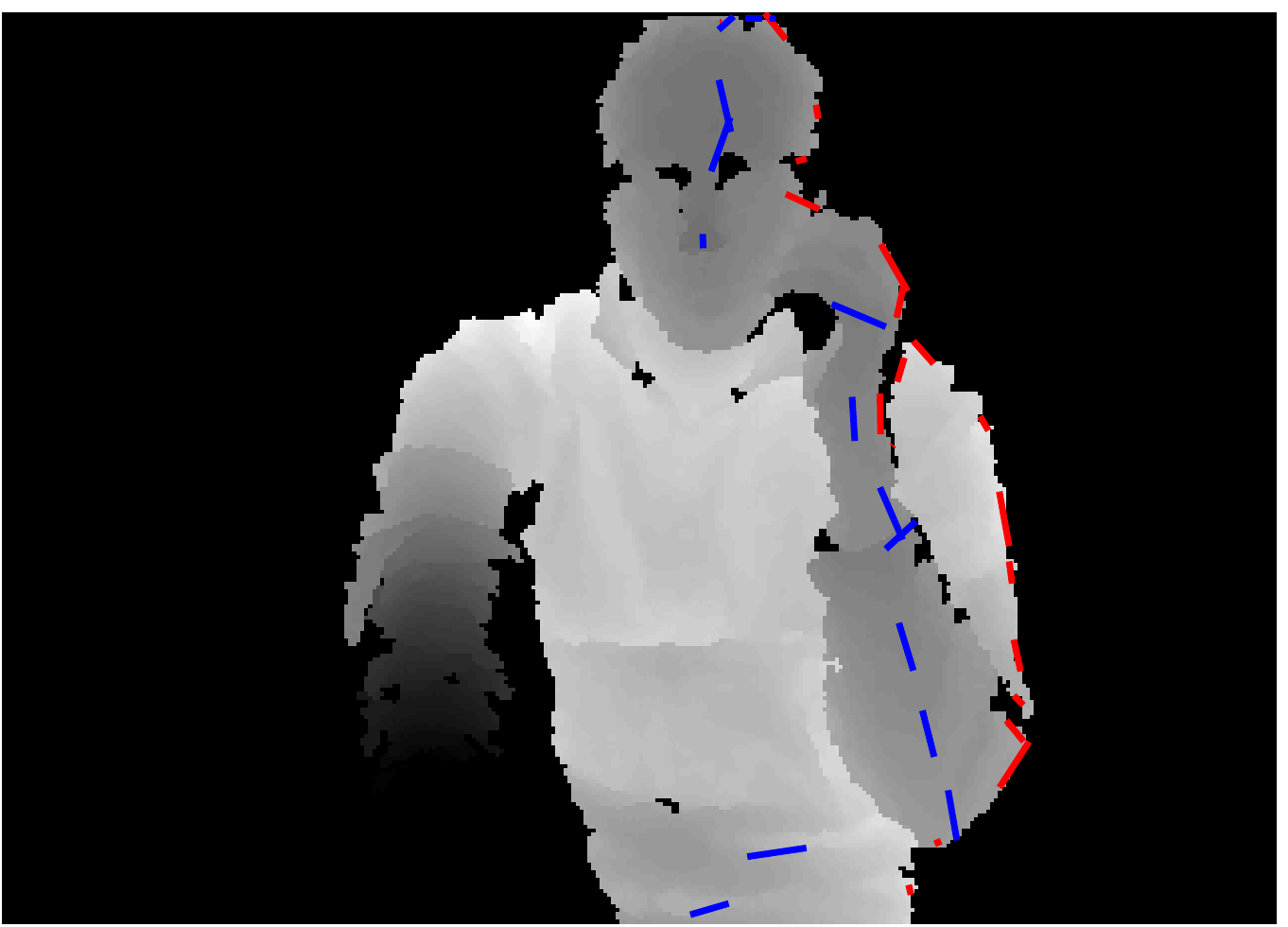} &
   \includegraphics[scale=0.25]{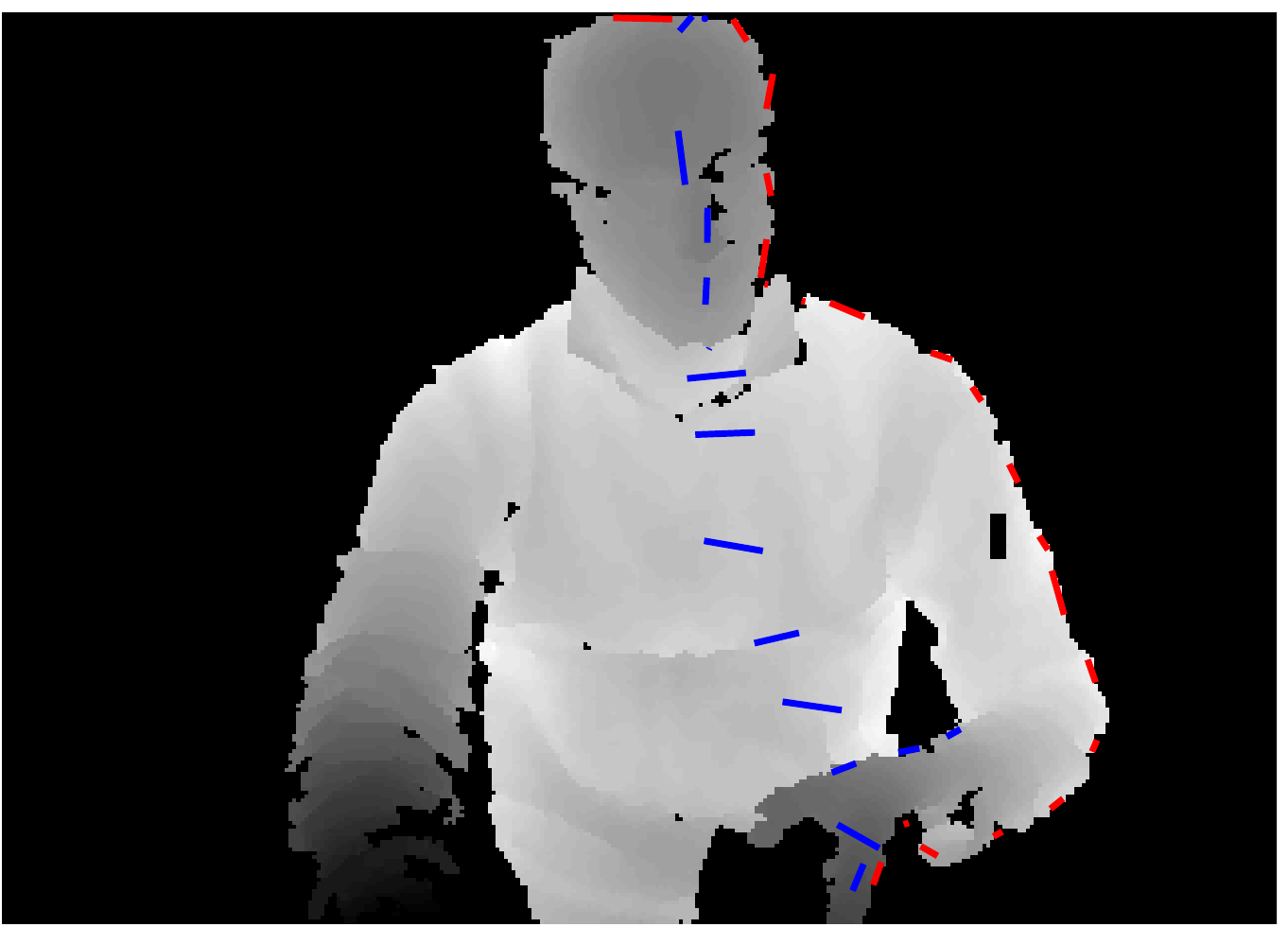} \\
   (a) Forward & (b) Right & (c) Up & (d) Down\\
\end{tabular}
\caption{First row: an example of frontal and associated profile views. Second row: examples of arm positions and associated features. Red dots are projections of point clouds' local orientations from the frontal view and blue dots from the profile view. As can be seen, profile view features are particularly useful in the case of up and down positions.}
\label{im:feature_examples}
\end{figure*}

Using only the right contour from the frontal view is not enough, as arm and especially forearm might not always be detected. To overcome this situation, we apply the aforementioned technique to what we call the profile view: each pixel of the depth map reprensents a point in 3D world coordinates. Suppose that the $\vec{X}$ and $\vec{Y}$ coordinates are the image pixel coordinates, and $\vec{Z}$ coordinate is the (depth) value of the pixel. Thus, the depth map corresponding to the frontal view is the projection of the 3D world points onto the  $(\vec{X},\vec{Y})$ plan. The profile view would then be the projection on the $(\vec{Y},\vec{Z})$ plan. Figure \ref{im:feature_examples} is an example of a segmented depth map and associated profile view. From this profile view, we apply contour detection and feature extraction just as described above. We now have 120 features. On figure \ref{im:feature_examples}, examples of features extracted for different poses are shown. Using the profile view makes possible the detection of the forearm even when the arm is in front of the body.\\

\subsubsection{Arm position output}
After feature extraction, the arm position module outputs an estimation of the arm position among four possible states: arm up, arm down, arm right, and arm forward. In that regard, we use a machine learning technique.

Among the 120 extracted features, some are discriminative, some are useless, and some might even be contradictory. Selecting or weighting the features is therefore necessary for good classification. We decided to use an AdaBoost \cite{freund1997decision}, which is a very appropriate tool in this regard. Nevertheless, AdaBoost only solves two-class problems, so we use a 1-vs-all approach: we train four sub-classifiers, each of them specialized for one class (i.e., using as positive examples some samples for a specific class, and as negative examples, some samples from any other class). The estimated position would be the one with highest value among the four sub-classifiers. The actual output of the module is not exactly the estimated position, but the score of each sub-classifier. Indeed, this four-values information is better-suited for the fusion stage.

\subsection{Eyes behavior}

\begin{figure*}
	  \centering
		\includegraphics[scale=0.55]{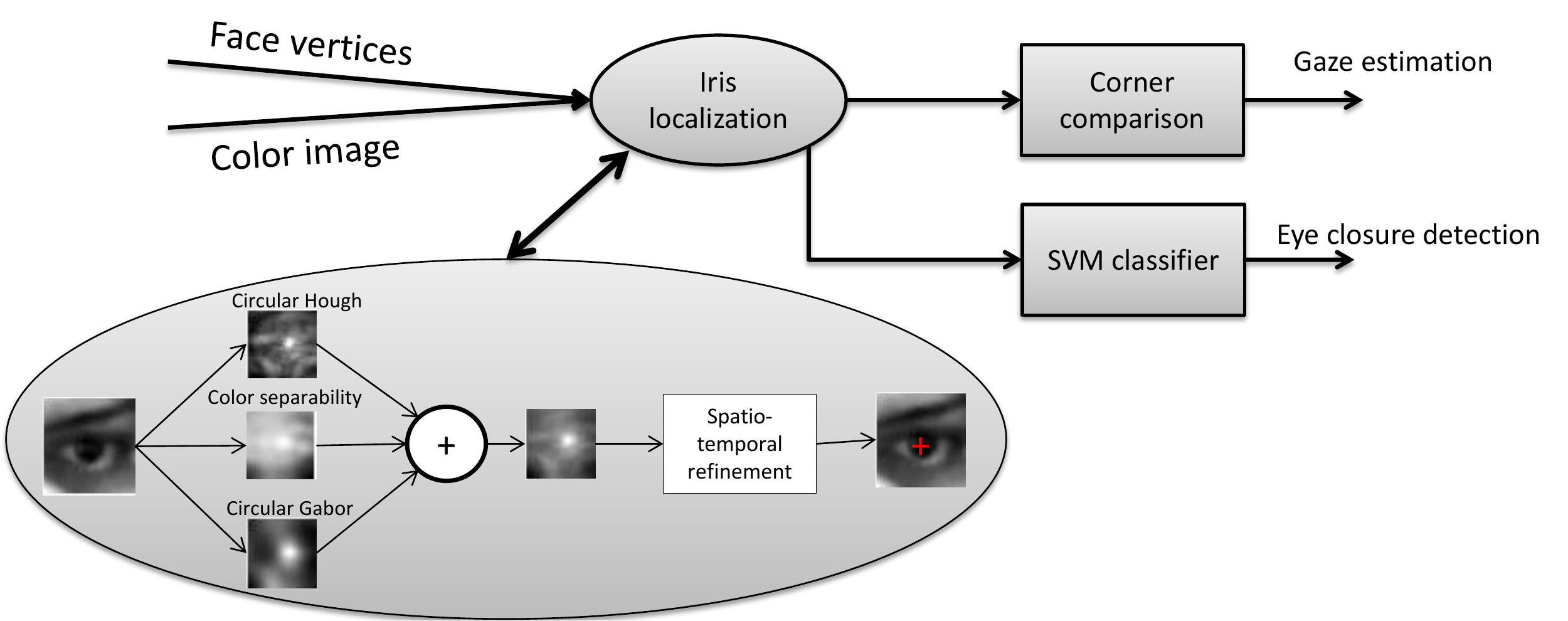}
    \caption{Main steps of the eye behavior module}
    \label{im:eye_flowchart}
\end{figure*}

Eye behavior is a very important feature for driver distraction, including both eye gaze and eye blinking. Therefore, this module aims to localize the iris of the driver and deduce his gaze position, while detecting whether the eye is open or closed. For this module, we only rely on the Kinect color stream. Figure \ref{im:eye_flowchart} shows a summary of the module.\\

\subsubsection{Iris localization}
Iris detection first relies on finding the eye corners positions. This is done using the face tracking algorithm provided by the Kinect SDK, further described in Section \ref{sec:facetracking}. We then use a robust iris detection method based on cost function maximization and spatio-temporal considerations. Given the eye corners locations, we isolate the eye by creating a squared patch with the same width as the eye corner distance. To make sure that our parameters are scale-invariant, we resize this square to a $60\times 60$ patch. Our cost function is based on the response of three different filters described below.\\

The first one is the circular Hough transform that has been widely used in iris detection problems, for example in \cite{1384533} or \cite{kawaguchi2003iris}. We choose a low edge detection threshold to make sure that the iris contour is detected in any situation, and we use a varying radius from six to nine pixels. Thus, we make sure that the iris will always be responsive to the filter.\\

The second filter was used in Zhang \textit{et al.} \cite{zhang2008new}, and relies on the circular Gabor filter. It provides significant impulse response to round objects, which is appropriate for iris detection. For that we convolve our eye template with the following Gabor kernel:

\begin{equation}
G(x,y) = g(x,y)exp(2i\pi F \sqrt{x^2+y^2})
\end{equation}

where $F$ is the radial frequency, set to 0.0884 and $g$ is the Gaussian envelope, defined as:

\begin{equation}
g(x,y) = \frac{1}{\sqrt{2\pi \sigma^2}}exp(-\frac{x^2+y^2}{2\sigma^2})
\end{equation}

with $\sigma$ the variance, set to 4.5 in our system.

\begin{figure}
	  \centering
		\includegraphics[scale=0.4]{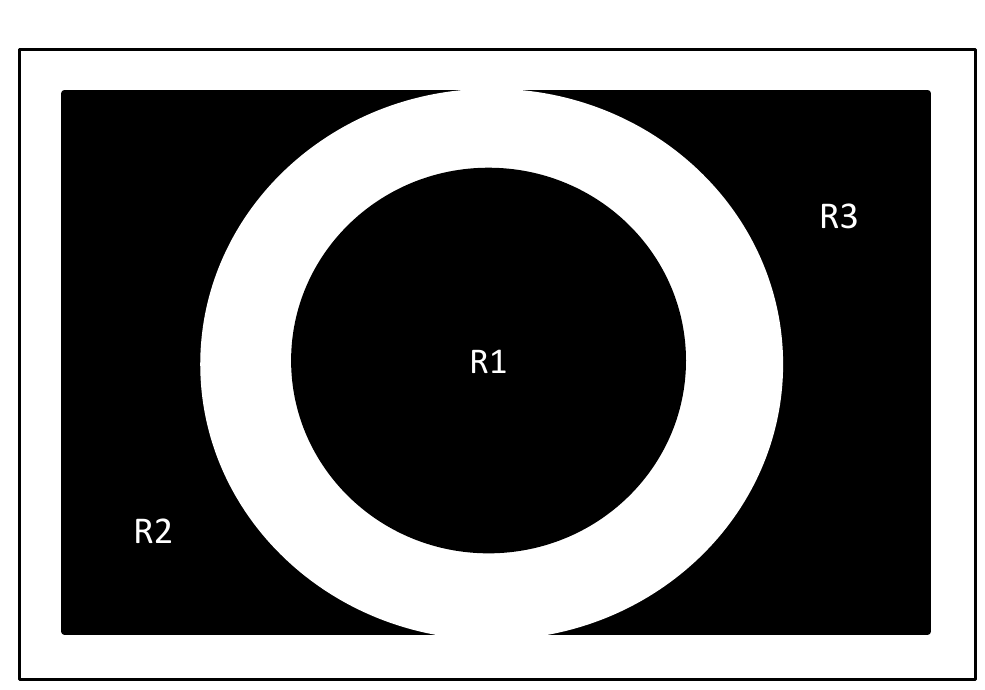}
    \caption{Masks $R_1$, $R_2$ and $R_3$ for separability measure. Dark areas represent pixel value 1 and white areas represent pixel value 0}
    \label{im:masks}
\end{figure}

The third filter is inspired by Kawaguchi \textit{et al.} \cite{kawaguchi2003iris}. It relies on the high intensity difference between the iris and its immediate neighborhood. For this, we convolve the eye template with the masks $R_1$, $R_2$, $R_3$ represented in Figure \ref{im:masks}. The radius of mask $R_1$ is 6, and for masks $R_2$ and $R_3$ is 15. We obtain three transformed eye templates, called $C_1$, $C_2$, and $C_3$. We combine them to obtain our separability response using the following formula:

\begin{equation}
S(x,y) = \frac{C_2(x,y)-C_1(x,y)}{C_1(x,y)} + \frac{C_3(x,y)-C_1(x,y)}{C_1(x,y)} 
\end{equation}

Last, we normalize our three filter responses, sum them up, and take the maximum value as our iris center estimate.

The detection is now quite accurate, but a few mistakes can be avoided and corrected using spatio-temporal information. We therefore use a simple spatio temporal consistency checking to improve the detection performance. For each template, we look at the predicted position of each four filter (including the sum of the three filters) and the previous estimated locations. We then apply simple rules based on distances and eye position to remove irrelevent detections.

\subsubsection{Gaze estimation}
Now that the iris is detected, we generate a feature set based on gaze estimation. Gaze can be estimated from a 3D model of the face determining the 3D-world orientation, and the iris location given the eyeballs' sizes and positions \cite{heyma-11}. We could use such an approach, but we need only a rough estimation of where the driver is looking, and statistical measures of the eye gaze distributions. For that reason, the features we extract are simply the relative position of the iris to the eyes' corners: from the face tracing, we extract the eye corner positions, and we generate the 4 dimensional feature vector as follows:

\begin{equation}
\begin{pmatrix} x_{l}\\y_{l}\\x_r\\y_r\end{pmatrix} = 
	\begin{pmatrix} \frac{X_l-C_l^{(l)}}{C_l^{(l)}-C_{l}^{(r)}} \\ \\ \frac{X_r-C_r^{(l)}}{C_r^{(l)}-C_{r}^{(r)}} \end{pmatrix}
\end{equation}

with $X_l$,$X_r$ the left (right) iris position and $C_i^{j}$ ($i,j \in {left, right}$) the $j$ corner of the $i$ eye.

\subsubsection{Eye closure detection}
We use an additional feature which determines whether the driver's irises are visible (eyes open) or not (eyes closed). This will be helpful when taking into account the amount of time the driver is not looking at the road and the potential danger this represents. A simple yet efficient approach for this is to construct a database of open and closed eyes, and to apply template matching or classification techniques \cite{Choi2011},\cite{arai2011comparative}, \cite{grauman2003communication}. We use a simple classification approach. When pupil position is estimated, we create a small iris template, centered at the iris location. We normalize the grayscale to make it more illumination-insensitive. We create a subset of around 2000 eyes templates, and we manually label each of them as whether it corresponds to an open eye with visible pupil, or a closed eye (or non-visible pupil). Using this dataset, we train an SVM classifier using an RBF kernel with $\sigma$ = 13. Last, we use the SVM for each session, and we add to the output of the module the SVM score (not the output label) for each eye and each frame.\\

\begin{figure*}
\centering
\begin{tabular}{cccc}
	 \includegraphics[scale=0.35]{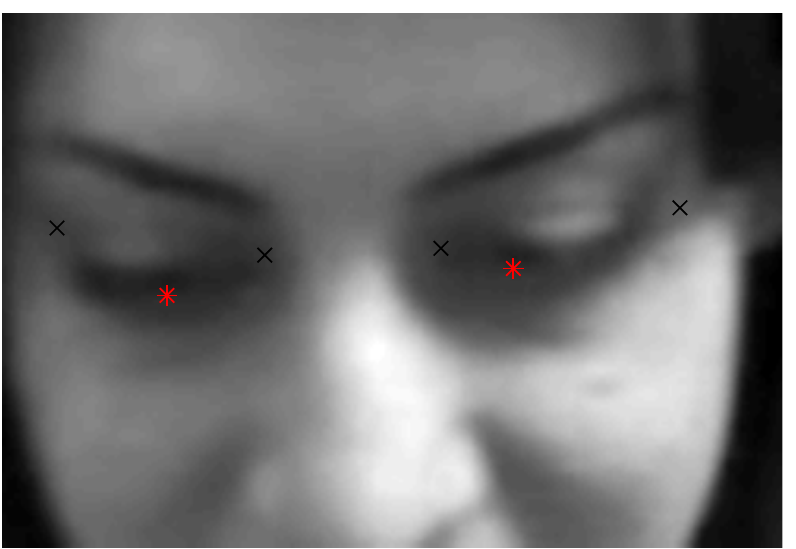} &
	 \includegraphics[scale=0.35]{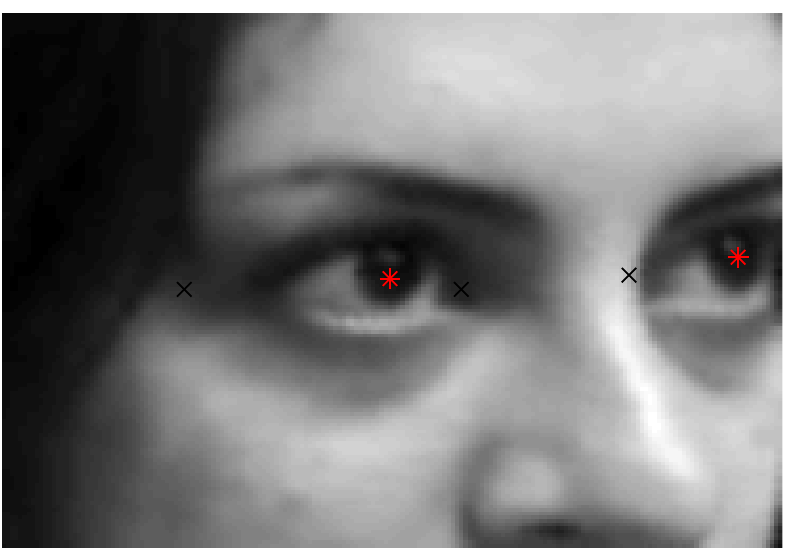} &
   \includegraphics[scale=0.35]{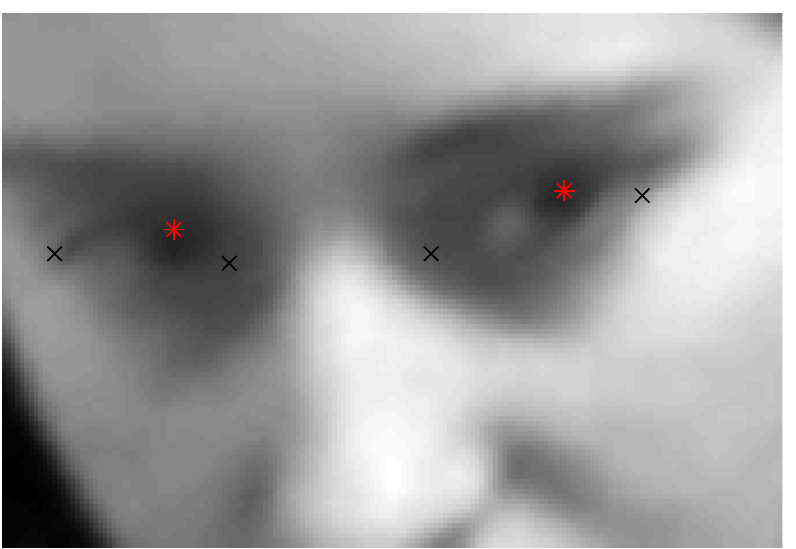}& 
	 \includegraphics[scale=0.35]{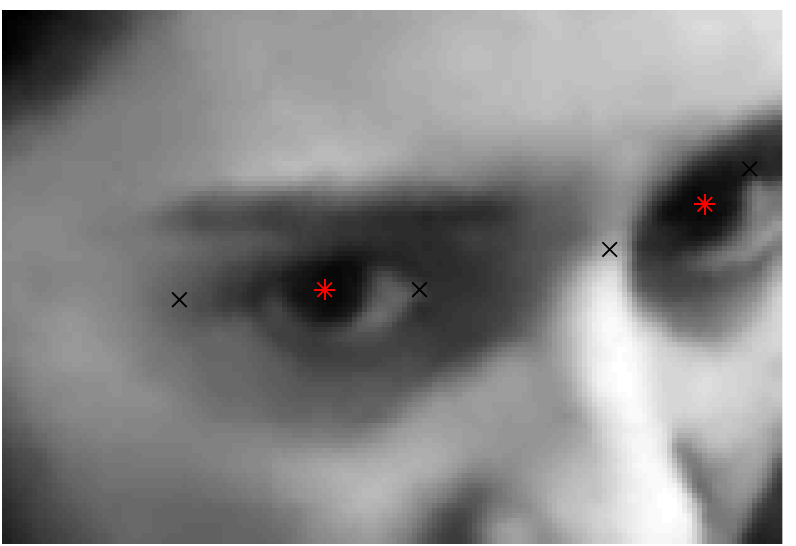} \\
   \includegraphics[scale=0.35]{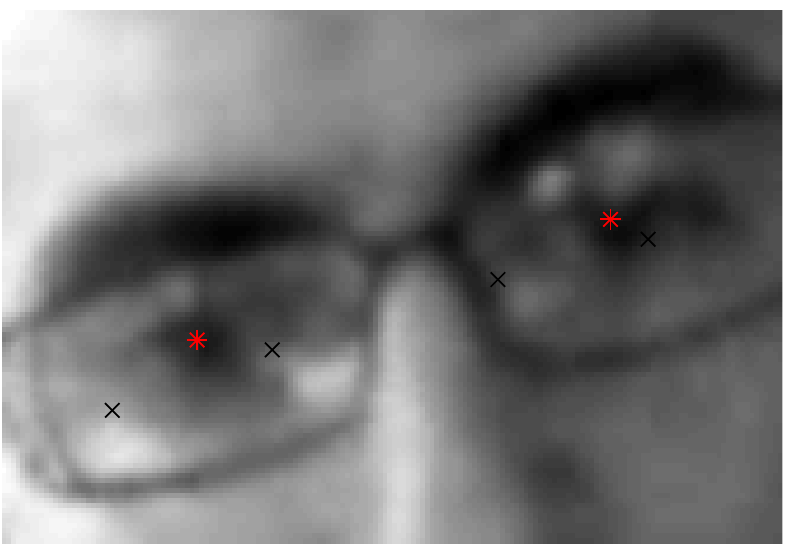} &
   \includegraphics[scale=0.35]{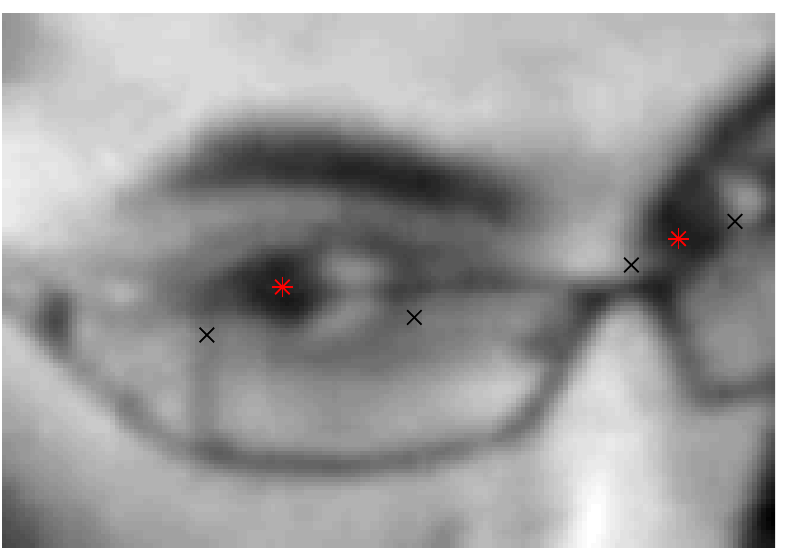} &
   \includegraphics[scale=0.35]{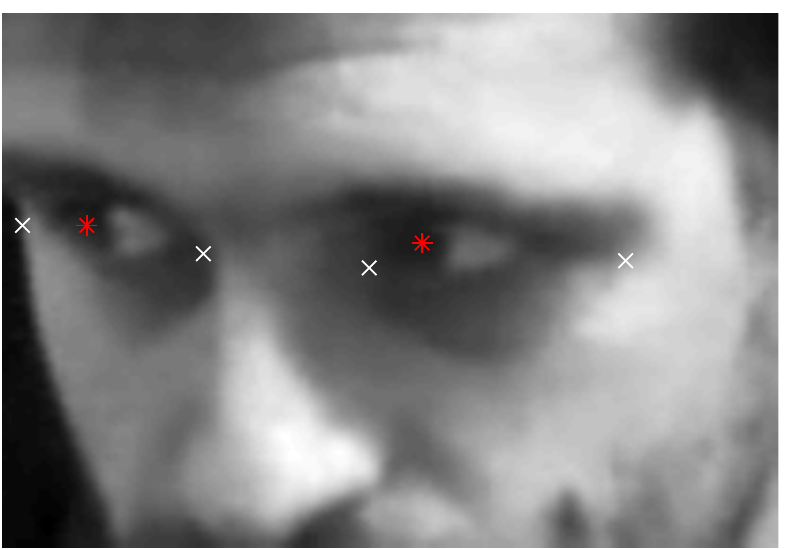} &
   \includegraphics[scale=0.35]{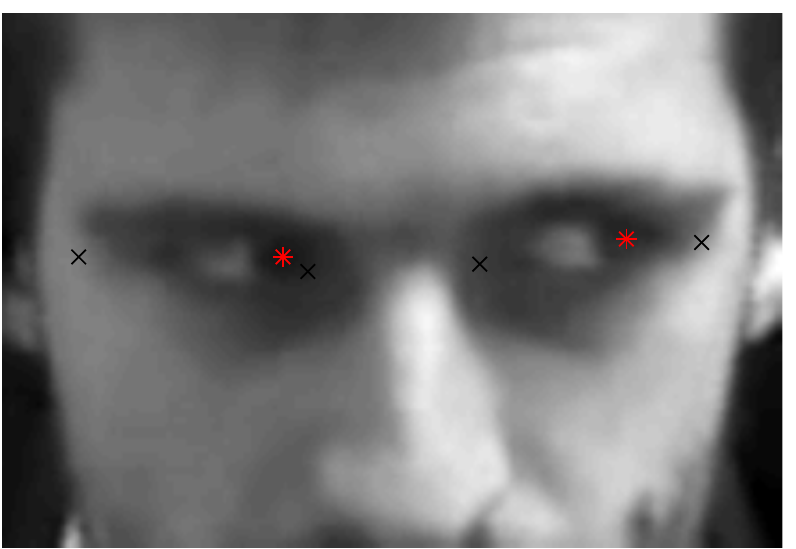} \\

\end{tabular}
\caption{Examples of iris detection involving various poses, expressions, and image qualities.}
\label{im:iris_detection_result}
\end{figure*}

Figure \ref{im:iris_detection_result} shows examples of pupil detection to qualitatively illustrate the performance of our detector. 

\subsection{Head pose and facial expression}
\label{sec:facetracking}
The Kinect SDK provides very useful features and cutting-edge algorithms to help developers. The face tracking algorithm \cite{faceTracking} is one of them, and reveals itself to be particularity useful for our work. It uses cooperative fusion of the depth map and the color image to  first estimate the root of the head, and then provides a robust and accurate face model. It relies on the active appearance model \cite{edwards1998interpreting} extended to 3D coordinates. The raw output of the face tracking is a set of 100 vertices and 121 triangles forming a mesh of the face. The face was tracked in most situations, but failed in case of rapid and significant face rotations, or when occluded by an object (typically when drinking). A few seconds were also required most of the time to correctly initialize the face model and fit it to the face. In the case where face tracking was not providing any output, the eye behavior module was deactivated as no eye location could be found.\\

Upper level information is also available from the face tracking, making the output much more meaningful and useful for our tasks. We use that high level information as the output of two of our modules.

\subsubsection{Face orientation} Based on 3D vertices coordinates, face tracking can provide head orientation angles and head center 3D position. For our work, we extract only the head orientation, namely the pitch, roll, and yaw angles, which are values between -180 and 180 degrees. The position depended too much on the driver's height and did not help in the classification task.

\subsubsection{Facial expression} Face tracking also provides six animation units (AUs) based on the definition of the Candide3 model \cite{ahlberg2001candide}. AUs are expressed as coefficients and represent how strongly distorted features of the face are. We extract only mouth-related AUs: upper lip raiser (AU10), jaw lowerer (AU26/27), lip stretcher (AU20), and lip corner depressor (AU13/15). Other AUs are eyebrows-related and did not help in the recognition task. More detailed information about AUs provided by the face tracker can be found on the Microsoft website \cite{faceTracking}. The output of the module is a set of 6 AUs extracted from the face model (if any).

\begin{figure*}
\centering
\begin{tabular}{ccc}
	 \includegraphics[scale=0.22]{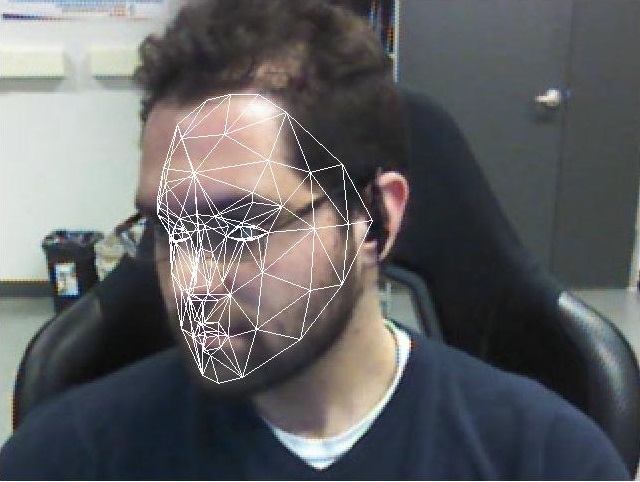} &
	 \includegraphics[scale=0.22]{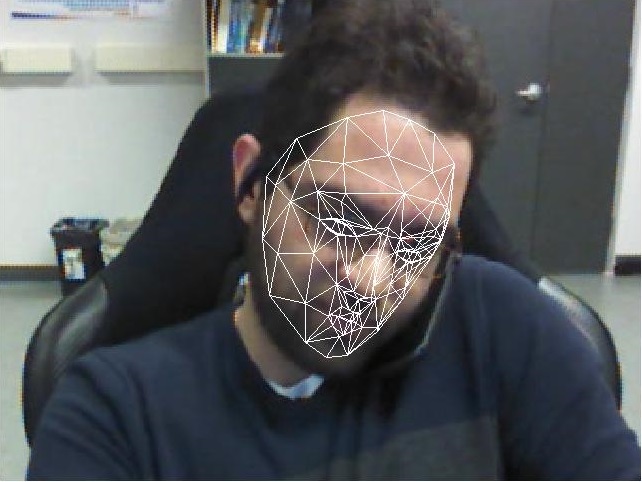} &
   \includegraphics[scale=0.22]{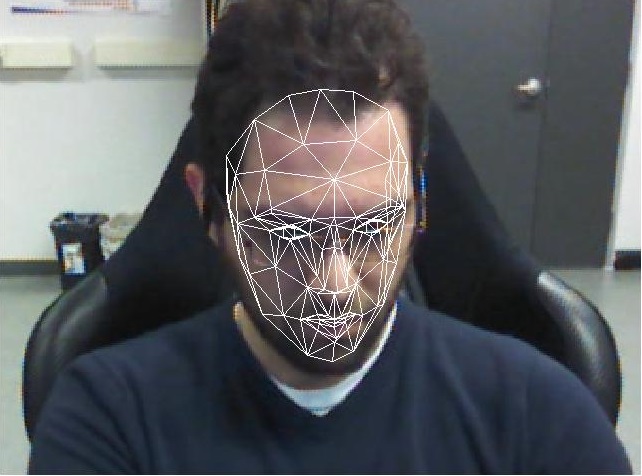} \\
    (a) & (b) & (c) \\
   \includegraphics[scale=0.22]{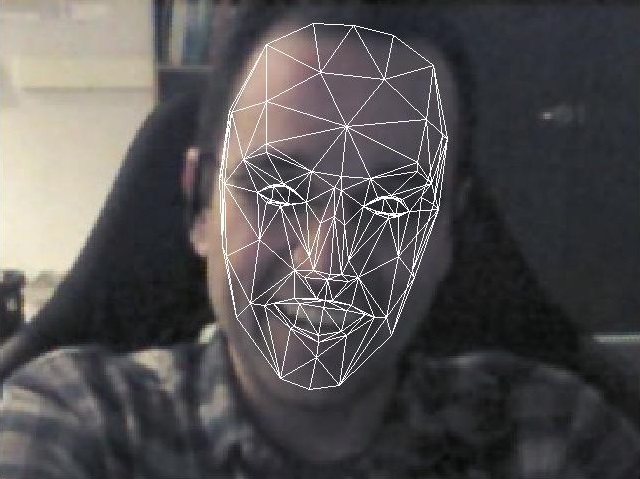} &
	 \includegraphics[scale=0.22]{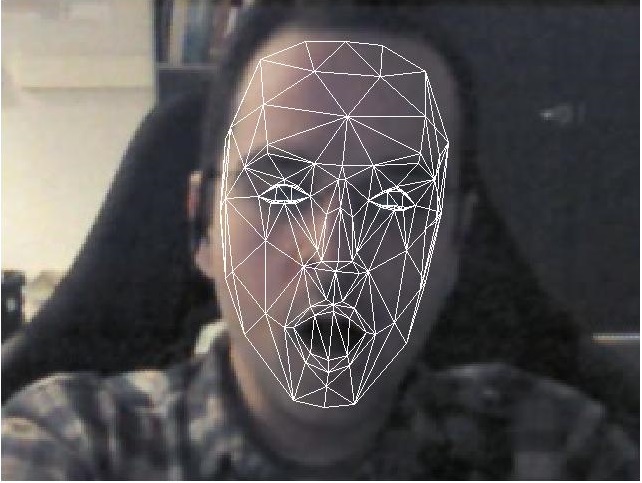} &
   \includegraphics[scale=0.22]{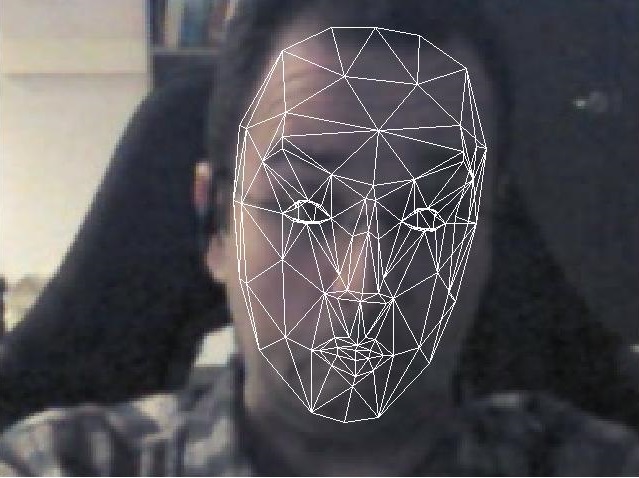} \\
   (d) & (e) & (f) \\
\end{tabular}
\caption{Examples of face tracking under different face poses (a),(b),(c) and facial expressions (d), (e), (f)}
\label{im:faceTracking}
\end{figure*}

Figure \ref{im:faceTracking} shows sample images and associated face tracking results. As can be seen, tracking is efficient in a number of situations.

\section{Fusion}
\label{sec:fusion}

In this section, we explain how to merge the aforementioned module outputs, and deduce the type of distractions taken by the driver.

Detecting distraction can be assimilated to a gesture recognition a gesture recognition problem. As gesture recognition is getting more and more popular in the fields of entertainment and security, in-vehicle applications also exists, for example for automatic commands \cite{westeyn2003georgia}. However,  our goal is slightly different and somehow more difficult in our case, as the driver is not cooperative when taking the action. Plus, there is no single way to accomplish an action which might be unexpectedly interrupted because of road constraints. Even for a human being, isolating and recognizing actions inside a car is not always obvious.

\subsection{Dataset}
A set of video sequences was collected on our experimental platform. Five distractive tasks were recorded and manually labelled for training and evaluation. Features were extracted based on the modules of Section \ref{sec:modules}, on a frame by frame basis. They were then concatenated such that a seventeen-features vector was representing each frame. More about the conditions of the experiments has been described in Section \ref{sec:experimental_results}.

\subsection{Feature extraction and temporal smoothing}
\label{sec:smooth}
Classification can be achieved using each frame independently, but a much better performance was obtained using temporal considerations. For this, we applied a running median filter, using a hundred-sample sliding window. This corresponds to a window length of approximately six seconds, which means that the beginning of an action was predicted with a delay of three seconds. We also computed the running standard deviation within the hundred-samples window, providing information about the temporal variation of each feature. The standard deviation information has shown better performance and stability compared to regular speed and acceleration features ($\delta$ and $\delta^2$ features). As a result, we obtained thirty-four features per sample. We can now train a classifier with a portion of this dataset to obtain a distraction recognition system.

\subsection{AdaBoost classifier}
Depending on the type of distraction we want to identify, the extracted features are either strongly discriminative or unsignificant. In that regard, AdaBoost can provide an appropriate solution. Briefly, AdaBoost (for Adaptive Boosting) is able to turn a set of T weak classifiers into a stronger one, by linearly combining each of them in an optimal way.

\begin{equation}
H(x) = sign(\sum_{t = 1}^T{\alpha_th_t(x)})
\end{equation}

At each iteration ($T$ in total), a weak classifier $h_t$ is selected from among a family of classifiers, namely, the one minimizing the weighted error rate. The weights and $\alpha_t$ are updated based on the minimum error rate. The weak classifiers we use are decision trees, which, by definition, are trained to select the most representative features maximizing the information gain.

As distraction identification is a multi-class classification problem, we used a 1-vs-all approach. Each class was trained using a simple real AdaBoost, initialized with a decision tree of depth four and 300 iterations.
The temporal aspect of the dataset can also help refining the frame-by-frame classification; therefore, we used an additional temporal filtering: we first classified each sample independently, and we replaced each estimated $S$ output by the mode (most frequent output) over a hundred-samples sliding window centered at $S$. This way, isolated misclassifications were removed.

\subsection{Hidden Markov Model }
Because of the temporal aspect of our dataset, stochastic models can be applied. Similar to speech recognition tasks, gesture recognition can be successfully achieved using Hidden Markov models (HMMs) \cite{lee1999hmm}, conditional random fields \cite{wang2006hidden}, or recursive types of classifiers, making the classification time-dependent \cite{murakami1991gesture}.
We chose to use HMMs, which provide a high-level of flexibility for modeling the structure of an observation sequence. HMMs allow recognition of the hidden state sequence by the dependency of the observation sequence on the hidden states and the topology of the state transitions. It is now acknowledged that the use of HMM is fundamental in temporal pattern recognition. 
We have developed our system using the HMM Toolkit (HTK) \cite{young2002htk}. The labels associated with each samples were the five possible states of distraction described in Section 

We train a different Markov model for each class, and used the Viterbi algorithm to decide which state each sample belongs to. We tried several configurations, varying the number of hidden states (from five to twenty) and the type of data (raw or smoothed, with or without $\delta$ and $\delta^2$ features).

\section{Experimental results}
\label{sec:experimental_results}
In this section, we demonstrate the efficiency of our system. First, a description of the experimental setup and dataset acquisition is described, then we evaluate the distraction detection and recognition using AdaBoost and Hidden Markov models.

\subsection{Experimental setup and data collection}

The data collection was obtained using a driving simulator. The simulator is equipped with monitors, steering wheel and driving software City Car driving \cite{CCD}. We used a Kinect sensor to record color video, depth map and audio. Steering wheel and pedal position as well as hear rate were also recorded. Last the entire driving session was captured based on a 1 fps screenshot of the monitors. In this paper, the only sensor used is the Kinect, but the other sensors will be used in a future work.\\
A total of 8 drivers were asked to participate in the study. Drivers
were either men of women, from different countries, either wearing
glasses or not with age varying from 24 to 40 years old. They were
all experimented drivers, but using their car at varying frequencies.
Each driver was recorded during four 15 minutes sessions. Two sessions
were in the early morning, when driver was awake and alert, and two
sessions in the late evening, at the end of a full business day, when
the driver was tired. Each session was either on a highway with low
traffic, or in the city with higher traffic. During each session,
the driver had to follow a well-defined procedure involving several
tasks putting him into distracted situations and miming visual signs.
During a driving session, about half of the time was normal driving,
and the other half was distracted driving. 

More precisely, the drivers were asked to accomplish five distracting
tasks during each driving session, namely
\begin{itemize}
\item making a phone call 
\item drinking 
\item sending an SMS 
\item looking at an object inside the vehicle (either a map or adjusting
the radio) 
\item driving normally
\end{itemize}

As a result, we have obtained around eight hours of driving sequences, that were manually labelled for analysis. The dataset was evaluated using cross validation: The performance
on each driver was evaluated separately by taking the data related
to the driver as the testing set, and the other drivers as the training
set. A ground truth was manually labeled for each session and was
available each second. The performance of the whole system was measured
by the total average accuracy, calculated according to the following
formula:

\begin{equation}
Total\, Average\, Accuracy=\frac{Number\, of\, correct\, decisions}{Total\, number\, of\, frames}\label{eq:Total_Average_Accuracy}
\end{equation}

\subsection{Results}

A first approach for classification is to use a time-independent classifier, and add temporal refinement to increase frame by-frame-accuracy. For practical reasons, we use an AdaBoost classifier again using a 1-vs-all approach. Each class is trained using a simple real AdaBoost initialized with a decision tree of depth four and 300 iterations.

We evaluate the action recognition using AdaBoost and HMM classifiers. For each driver, we evaluate the distraction recognition capacity by training a classifier using all driver sessions except the driver to be evaluated, and testing using all sessions involving this driver. For AdaBoost, the best accuracy was obtained using a Real AdaBoost algorithm with initialization based on a decision tree of depth four and 300 iterations. For HMM, the optimal parameters were a ten-states automaton with a single Gaussian mixture for modeling each node, and using the smoothed data described in Section \ref{sec:smooth} rather than the original raw data. No significant improvement was found when using $\delta$ and $\delta^2$ features.\\

Table \ref{table_overall} presents the overall accuracy of the AdaBoost and HMM classifiers for each driver. We provide the accuracy for the five classes and the accuracy for distraction detection only: in this case, we have merged all the classes involving distraction into a single class, and compared it with the normal driving class. Average accuracies are quite close between AdaBoost and HMM (85.05\% and 84.78\%), but the results for each driver may vary. More specifically, HMM is outperforms AdaBoost for most of the drivers, but for a few drivers, HMM performs significantly worse than AdaBoost. This may be due to instabilities related to the high dimensionality of the features and the number of states compared to the size of our training set. Increasing the number of drivers might solve this issue. Also, one might be surprised that HMM does not perform as superbly as it does for the gesture recognition task. Again, this is because the driver is not cooperative in this case, and there is no single way to accomplish a distractive task. For example, when drinking, a driver can put the container down between each swallow, or just keep it in the hand. When phoning, the driver can place the phone between his head and his shoulder, or keep it in hand. Moreover, actions can be interrupted suddenly, because the driver needs both hands on the wheel to turn, or greater focus on the road to avoid accidents. All these constraints make the actions less similar in time, and significantly limits the HMM performance. Using a bigger dataset might eventually improve the classification results. As regards distraction-only accuracy (89.84\% and 89.64\%), results suggest that our system can successfully detect whether a driver is actually distracted or not.\\

\begin{table*}
\caption{Accuracies of AdaBoost and HMM classifiers for distraction recognition (5 classes) and distraction detection (2 classes)}
\label{table_overall}
\centering
\begin{tabular}{|>{\centering\arraybackslash}p{2.2cm}|>{\centering\arraybackslash}p{2.2cm}|>{\centering\arraybackslash}p{2.2cm}|>{\centering\arraybackslash}p{2.2cm}|>{\centering\arraybackslash}p{2.2cm}|}
\hline
 & \multicolumn{2}{c|}{\bf AdaBoost} & \multicolumn{2}{c|}{\bf HMM} \\
\hline 
Subject  				 & Distraction recognition & Distraction detection & Distraction recognition & Distraction detection \\
\hline
1  			 & 89.36		   & 91.44 			 			& 84.00 			& 92.31 \\ 
\hline
2 			 & 86.02		   & 90.05 			 			& 87.41 			& 91.16 \\ 
\hline
3 		 & 87.38		   & 90.95 			 			& 90.19 			& 96.41 \\ 
\hline
4 			 & 85.48		   & 94.72 			 			& 81.85 			& 84.65 \\ 
\hline
5 		 & 81.14		   & 85.05 			 			& 83.68 			& 83.82 \\ 
\hline
6 	 		 & 80.94		   & 87.7 			 			& 81.52 			& 89.53 \\ 
\hline
\bf{Average} 		 & \bf{85.05}  & \bf{89.84} 	 		& \bf{84.78}	& \bf{89.64} \\
\hline
\end{tabular}
\end{table*}

Another test we did was to train the AdaBoost classifier with each module separately and evaluate their inference capacity (See table \ref{tab:accuPerFeatures}). We found that features related with arm position were by far the most discriminative, probably because normal driving position was easy to detect, and represented almost 40\% of a driving session. Other features were also useful, but provided efficient recognition only for specific actions. For example, we found that face orientation was very discriminative for drinking and text messaging, face expression helped a lot differentiating phone call and drinking and eyes behaviour was efficient for text messaging and normal driving. AdaBoost was an appropriate choice of classifier, as decision trees as weak classifiers were doing feature selection for each type of action.

\begin{table*}
\centering
\caption{Accuracies per features}
\label{tab:accuPerFeatures}
\begin{tabular}{|>{\centering\arraybackslash}p{2cm}|>{\centering\arraybackslash}p{2cm}|>{\centering\arraybackslash}p{2cm}|>{\centering\arraybackslash}p{2cm}|>{\centering\arraybackslash}p{2cm}|>{\centering\arraybackslash}p{2cm}|}
\hline
Subject  				 & Arm   		& Orientation  & Expression  & Eyes  & All \\
\hline
1 			 & 76.49 		& 55.9 			 	 & 54.31 			 & 65.7  & 89.36 \\ 
\hline
2 			 & 70.04 		& 56.35 		   & 55.45 			 & 61.72 & 86.02\\ 
\hline
3   		 & 85.99 		& 66.8 			 	 & 56.48 			 & 38.6  & 87.38 \\ 
\hline
4 			 & 69.80 		& 63.63 			 & 55.09 			 & 59.95 & 85.48\\ 
\hline
5 		 & 74.55 		& 66.6 			 	 & 56.30 			 & 59.06 & 81.14\\ 
\hline
6 		 & 81.01 		& 77.87 			 & 64.25 			 & 63.85 & 80.94 \\ 
\hline
\end{tabular}
\end{table*}

The overall accuracy is a good indicator of system performance, but it does not say how each class is correctly detected. Table \ref{tab:each_class} provides a few classification metrics for each class, based on the average of each driver performance. For each class, extremelly high accuracy is due to an unbalanced testing set. Indeed, a single action (except normal driving) represents between 10\% and 20\% of the entire sequence. Therefore, recall and precision measures are better indicators of the classification capacity. From the table, it is clear that phone call and normal driving were quite successfully detected. Drinking was a little behind, mainly because the action was sometimes very fast (the driver just swallowed for a few seconds and put the drink down) and attenuated by temporal smoothing. The worst performance was for object distraction, probably because this action required neither huge visual nor cognitive attention. In that regard, the action was pretty similar to normal driving and therefore hard to detect, even for a human being. 

\begin{figure*}
		\includegraphics[scale = 0.35]{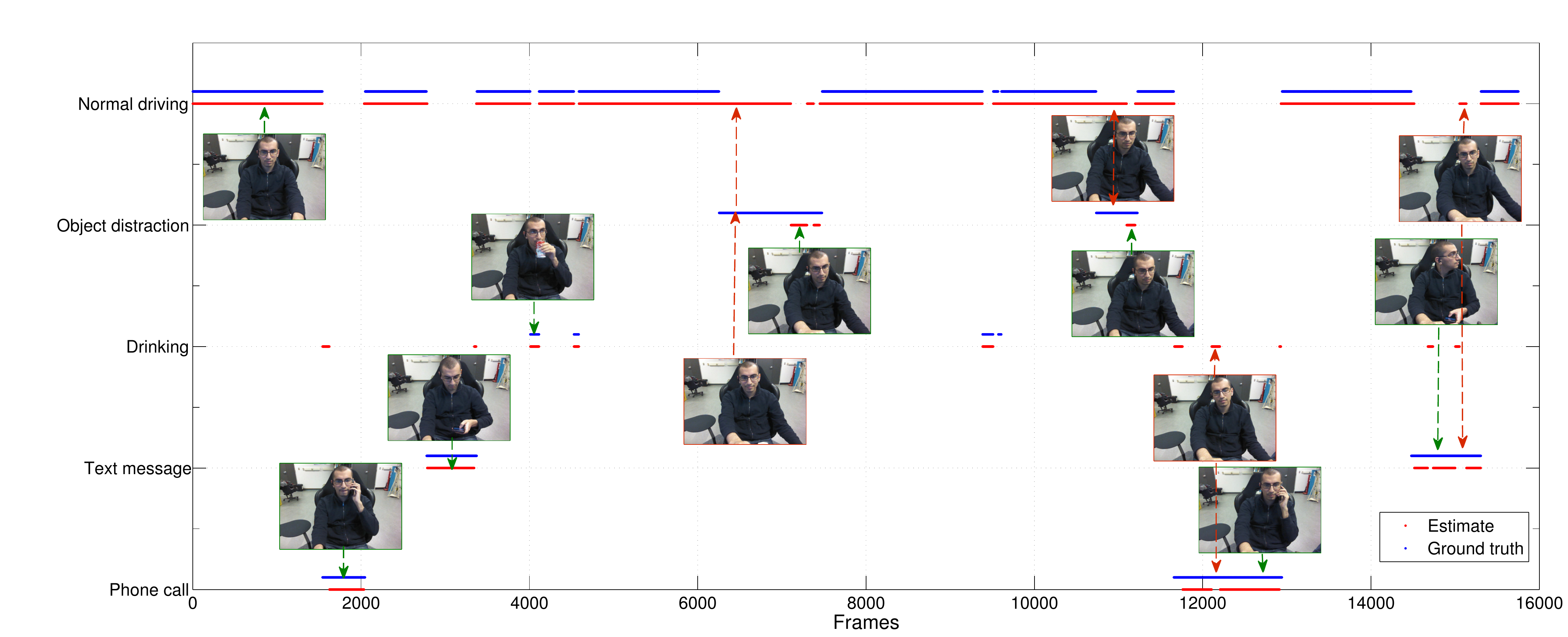}
    \caption{Results of action recognition for a given sequence. Ground truth (red) and estimated actions (blue) are displayed for each frame (x axis). Sample frames are displayed to illustrate the correct and wrong classification results for some actions. Green frames means correctly detected, red frame means wrong detection}
    \label{im:result_classification}
\end{figure*}

In order to get more insight about those results, figure \ref{im:result_classification} displays the frame by frame classification for a given sequence. Ground truth is the blue lines, and estimated class is the red one. In this example, phone call and text message are accurately detected, drinking comes with a few false positives and object distraction is often considered as normal driving. We have added a few frames to provide a better visualization of why detection was successful or not. Correct detections are represented with green frames and arrow, whereas false detections are in red. The drinking false positives are often due to strong arm movements (when changing gear during phone call for example). We believe that a better temporal analysis could remove that type of false positive detections. For object distraction misclassification, the sample frames show that the driver does not look extremely distracted, and it can be hard to say if he is actually adjusting the radio or just having his hand on the gear lever. Fortunately object distraction is the less demanding distraction and therefore the less dangerous, making the misclassification in that case less critical.

\begin{table*}
\centering
\caption{Classification measures for each class}
\label{tab:each_class}
\begin{tabular}{|>{\centering\arraybackslash}p{2cm}|>{\centering\arraybackslash}p{2cm}|>{\centering\arraybackslash}p{2cm}|>{\centering\arraybackslash}p{2cm}|>{\centering\arraybackslash}p{2cm}|>{\centering\arraybackslash}p{2cm}|}
\hline
Subject  		 & Phone call & Text message & Drinking  & Object distraction  & Normal driving \\
\hline
Accuracy  	 & 95.54 			& 96.24 			 & 95.55 	 & 92.79  & 89.98 \\
\hline
Sensitivity	/recall & 80.56 & 73.63 	   & 86.14 	 & 24.78 & 96.00\\
\hline
Specificity  & 97.90 			& 98.65 			 & 96.19 	 & 98.91  & 87.38 \\
\hline
Precision 	 & 87.51 			& 89.22 			 & 51.45 	 & 68.28 & 85.48\\
\hline
f measure    & 81.71 			& 72.04			 	 & 60.79 	 & 26.30 & 81.14\\
\hline
g-means	     & 87.98			& 78.32 			 & 90.91 	 & 40.16 & 80.94 \\
\hline
\end{tabular}
\end{table*}

\section{Conclusion and future work}
\label{sec:conclusion}
We have successfully built an inattention detection and recognition tool based on Kinect sensor, computer vision and machine learning. Based on four modules extracting data from arm position, face orientation, face expression and eye behaviour, we have constructed two types of classifiers to perform inattention action recognition. Based on data collected with a driving simulator, we were able to evaluate our work, and results show that our method is accurate and might be used in a real car.\\

Compared to existing approaches aiming to detect inattention or provide a level of inattention, our system outputs higher level information, more suitable for context aware human-machine interaction. Not only can it be used for immediate driver safety, but also for long term statistics about the driver habits, or for inter vehicles communications systems. Moreover, the different modules we have constructed are extremely flexible and could be used for other type of statistics computation and inference.\\

Future work directions are as follows: first, we believe that action recognition could be improved using more temporal information. For example, drinking and phone call are sometimes mixed up and alternating in successive frames, whereas it is very unlikely in practice that a driver is doing both at the same time. Such mistakes could be avoided. Next, the modules we have designed could allow fatigue detection using PERCLOS, nodding and yawning frequencies for example. Therefore, efforts will be put on fatigue detection in future work. We also plan to use background and environment information to assess the level of risk on the road. We might use a Bayesian network and later on dynamic Bayesian network in order to fuse multiple sources related to the driver (age, driving experience, fatigue) and the environment (time of the day, road type, outside traffic, vehicle speed) in addition to distraction type to assess the level of danger the driver is exposed to. Moreover, during the recorded sessions, we also have used additional sensors such as microphone, heart rate monitor or steering wheel signals. We will work on integrating those signals to the system. Last, we also plan to record more driving sessions, involving additional actions, per say interactions and chatting with other passengers.




\begin{thebibliography}{10}
\providecommand{\url}[1]{#1}
\csname url@samestyle\endcsname
\providecommand{\newblock}{\relax}
\providecommand{\bibinfo}[2]{#2}
\providecommand{\BIBentrySTDinterwordspacing}{\spaceskip=0pt\relax}
\providecommand{\BIBentryALTinterwordstretchfactor}{4}
\providecommand{\BIBentryALTinterwordspacing}{\spaceskip=\fontdimen2\font plus
\BIBentryALTinterwordstretchfactor\fontdimen3\font minus
  \fontdimen4\font\relax}
\providecommand{\BIBforeignlanguage}[2]{{%
\expandafter\ifx\csname l@#1\endcsname\relax
\typeout{** WARNING: IEEEtran.bst: No hyphenation pattern has been}%
\typeout{** loaded for the language `#1'. Using the pattern for}%
\typeout{** the default language instead.}%
\else
\language=\csname l@#1\endcsname
\fi
#2}}
\providecommand{\BIBdecl}{\relax}
\BIBdecl

\bibitem{distraction.gov}
NHTSA, ``Distraction.gov, official us government website for distracted
  driving,'' http://http://www.distraction.gov, 2013.

\bibitem{stutts2003driver}
J.~C. Stutts and W.~W. Hunter, ``Driver inattention, driver distraction and
  traffic crashes,'' \emph{ITE Journal}, vol.~73, no.~7, pp. 34--45, 2003.

\bibitem{strayer2004profiles}
D.~L. Strayer and F.~A. Drews, ``Profiles in driver distraction: Effects of
  cell phone conversations on younger and older drivers,'' \emph{Human
  factors}, vol.~46, no.~4, pp. 640--649, 2004.

\bibitem{blueprint}
N.~H. T.~S. Administration, ``Blueprint for ending distracted driving. dot hs
  811 629. u.s.'' Department of Transportation,Washington, DC., Tech. Rep.,
  2012.

\bibitem{Trafficsafetyfacts}
------, ``Traffic safety facts, research note dot hs 811 737. u.s.'' Department
  of Transportation,Washington, DC., Tech. Rep., 2013.

\bibitem{olson2009driver}
R.~L. Olson, R.~J. Hanowski, J.~S. Hickman, and J.~L. Bocanegra, ``Driver
  distraction in commercial vehicle operations,'' Tech. Rep., 2009.

\bibitem{williamson2005review}
A.~Williamson and T.~Chamberlain, ``Review of on-road driver fatigue monitoring
  devices,'' 2005.

\bibitem{young2007driver}
K.~Young, M.~Regan, and M.~Hammer, ``Driver distraction: A review of the
  literature,'' \emph{Distracted driving. Sydney, NSW: Australasian College of
  Road Safety}, pp. 379--405, 2007.

\bibitem{dong2011driver}
Y.~Dong, Z.~Hu, K.~Uchimura, and N.~Murayama, ``Driver inattention monitoring
  system for intelligent vehicles: A review,'' \emph{Intelligent Transportation
  Systems, IEEE Transactions on}, vol.~12, no.~2, pp. 596--614, 2011.

\bibitem{regan2011driver}
M.~A. Regan, C.~Hallett, and C.~P. Gordon, ``Driver distraction and driver
  inattention: Definition, relationship and taxonomy,'' \emph{Accident Analysis
  \& Prevention}, vol.~43, no.~5, pp. 1771--1781, 2011.

\bibitem{shiwu2011active}
L.~Shiwu, W.~Linhong, Y.~Zhifa, J.~Bingkui, Q.~Feiyan, and Y.~Zhongkai, ``An
  active driver fatigue identification technique using multiple physiological
  features,'' in \emph{Mechatronic Science, Electric Engineering and Computer
  (MEC), 2011 International Conference on}.\hskip 1em plus 0.5em minus
  0.4em\relax IEEE, 2011, pp. 733--737.

\bibitem{lal2002driver}
S.~K. Lal and A.~Craig, ``Driver fatigue: electroencephalography and
  psychological assessment,'' \emph{Psychophysiology}, vol.~39, no.~3, pp.
  313--321, 2002.

\bibitem{ranney2008driver}
T.~A. Ranney, ``Driver distraction: A review of the current
  state-of-knowledge,'' Tech. Rep., 2008.

\bibitem{Volvo}
M.~V. Car, ``Driver alert control,'' http://www.media.volvocars.com, 2007.

\bibitem{pilutti1997identification}
T.~Pilutti and A.~G. Ulsoy, ``Identification of driver state for lane-keeping
  tasks: experimental results,'' in \emph{American Control Conference, 1997.
  Proceedings of the 1997}, vol.~5.\hskip 1em plus 0.5em minus 0.4em\relax
  IEEE, 1997, pp. 3370--3374.

\bibitem{angell2006driver}
L.~Angell, J.~Auflick, P.~Austria, D.~Kochhar, L.~Tijerina, W.~Biever,
  T.~Diptiman, J.~Hogsett~Jr, and S.~Kiger, ``Driver workload metrics
  project,'' \emph{National Highway Traffic Safety Administration}, 2006.

\bibitem{dinges1998perclos}
D.~F. Dinges and R.~Grace, ``Perclos: A valid psychophysiological measure of
  alertness as assessed by psychomotor vigilance,'' \emph{Federal Highway
  Administration. Office of motor carriers, Tech. Rep. MCRT-98-006}, 1998.

\bibitem{ji2006probabilistic}
Q.~Ji, P.~Lan, and C.~Looney, ``A probabilistic framework for modeling and
  real-time monitoring human fatigue,'' \emph{Systems, Man and Cybernetics,
  Part A: Systems and Humans, IEEE Transactions on}, vol.~36, no.~5, pp.
  862--875, 2006.

\bibitem{Ji2004}
Q.~Ji, Z.~Zhu, and P.~Lan, ``{Real-Time Nonintrusive Monitoring and Prediction
  of Driver Fatigue},'' \emph{IEEE Transactions on Vehicular Technology},
  vol.~53, no.~4, pp. 1052--1068, 2004.

\bibitem{bergasa2008analysing}
L.~M. Bergasa, J.~M. Buenaposada, J.~Nuevo, P.~Jimenez, and L.~Baumela,
  ``Analysing driver's attention level using computer vision,'' in
  \emph{Intelligent Transportation Systems, 2008. ITSC 2008. 11th International
  IEEE Conference on}.\hskip 1em plus 0.5em minus 0.4em\relax IEEE, 2008, pp.
  1149--1154.

\bibitem{Bergasa2006}
L.~Bergasa, J.~Nuevo, M.~Sotelo, R.~Barea, and M.~Lopez, ``Real-time system for
  monitoring driver vigilance,'' \emph{Intelligent Transportation Systems, IEEE
  Transactions on}, vol.~7, no.~1, pp. 63 --77, 2006.

\bibitem{Senaratne2007}
R.~Senaratne, D.~Hardy, B.~Vanderaa, and S.~Halgamuge, ``Driver fatigue
  detection by fusing multiple cues,'' in \emph{Advances in Neural Networks 
  ISNN 2007}, ser. Lecture Notes in Computer Science, D. Liu, S.~Fei, Z.~Hou,
  H.~Zhang, and C.~Sun, Eds.\hskip 1em plus 0.5em minus 0.4em\relax Springer
  Berlin Heidelberg, 2007, vol. 4492, pp. 801--809.

\bibitem{smith2003determining}
P.~Smith, M.~Shah, and N.~da~Vitoria~Lobo, ``Determining driver visual
  attention with one camera,'' \emph{Intelligent Transportation Systems, IEEE
  Transactions on}, vol.~4, no.~4, pp. 205--218, 2003.

\bibitem{DOrazio2007}
T.~D'Orazio, M.~Leo, C.~Guaragnella, and a.~Distante, ``{A visual approach for
  driver inattention detection},'' \emph{Pattern Recognition}, vol.~40, no.~8,
  pp. 2341--2355, 2007.

\bibitem{craye2013multi}
C.~Craye and F.~Karray, ``Multi-distributions particle filter for eye tracking
  inside a vehicle,'' in \emph{Image Analysis and Recognition}.\hskip 1em plus
  0.5em minus 0.4em\relax Springer, 2013, pp. 407--416.

\bibitem{limulti2012}
L.~Li, K.~Werber, C.~F. Calvillo, K.~D. Dinh, A.~Guarde, and A.~K{\"o}nig,
  ``Multi-sensor soft-computing system for driver drowsiness detection,''
  \emph{Online conference on soft computing in industrial applications}, pp.
  1--10, 2012.

\bibitem{damousis2008fuzzy}
I.~G. Damousis and D.~Tzovaras, ``Fuzzy fusion of eyelid activity indicators
  for hypovigilance-related accident prediction,'' \emph{Intelligent
  Transportation Systems, IEEE Transactions on}, vol.~9, no.~3, pp. 491--500,
  2008.

\bibitem{eyeAlert}
EyeAlert, ``Distracted driving and fatigue sentinels,''
  http://www.eyealert.com/, 2012.

\bibitem{daza2011drowsiness}
I.~Daza, N.~Hernandez, L.~Bergasa, I.~Parra, J.~Yebes, M.~Gavilan, R.~Quintero,
  D.~Llorca, and M.~Sotelo, ``Drowsiness monitoring based on driver and driving
  data fusion,'' in \emph{Intelligent Transportation Systems (ITSC), 2011 14th
  International IEEE Conference on}.\hskip 1em plus 0.5em minus 0.4em\relax
  IEEE, 2011, pp. 1199--1204.

\bibitem{fletcher2005correlating}
L.~Fletcher, G.~Loy, N.~Barnes, and A.~Zelinsky, ``Correlating driver gaze with
  the road scene for driver assistance systems,'' \emph{Robotics and Autonomous
  Systems}, vol.~52, no.~1, pp. 71--84, 2005.

\bibitem{held20123d}
R.~Held, A.~Gupta, B.~Curless, and M.~Agrawala, ``3d puppetry: a kinect-based
  interface for 3d animation,'' in \emph{Proceedings of the 25th annual ACM
  symposium on User interface software and technology}.\hskip 1em plus 0.5em
  minus 0.4em\relax ACM, 2012, pp. 423--434.

\bibitem{PuppetParade}
T.~Watson and G.~Emily, ``Puppet parade - kinect arm tracker by design-io,''
  https://github.com/ofTheo/kinectArmTracker, 2012.

\bibitem{zhu2007constrained}
Y.~Zhu and K.~Fujimura, ``Constrained optimization for human pose estimation
  from depth sequences,'' in \emph{Computer Vision--ACCV 2007}.\hskip 1em plus
  0.5em minus 0.4em\relax Springer, 2007, pp. 408--418.

\bibitem{shotton2013real}
J.~Shotton, T.~Sharp, A.~Kipman, A.~Fitzgibbon, M.~Finocchio, A.~Blake,
  M.~Cook, and R.~Moore, ``Real-time human pose recognition in parts from
  single depth images,'' \emph{Communications of the ACM}, vol.~56, no.~1, pp.
  116--124, 2013.

\bibitem{lorensen1987marching}
W.~E. Lorensen and H.~E. Cline, ``Marching cubes: A high resolution 3d surface
  construction algorithm,'' in \emph{ACM Siggraph Computer Graphics}, vol.~21,
  no.~4.\hskip 1em plus 0.5em minus 0.4em\relax ACM, 1987, pp. 163--169.

\bibitem{freund1997decision}
Y.~Freund and R.~E. Schapire, ``A decision-theoretic generalization of on-line
  learning and an application to boosting,'' \emph{Journal of computer and
  system sciences}, vol.~55, no.~1, pp. 119--139, 1997.

\bibitem{1384533}
Q.-C. Tian, Q.~Pan, Y.-M. Cheng, and Q.-X. Gao, ``Fast algorithm and
  application of hough transform in iris segmentation,'' in \emph{Machine
  Learning and Cybernetics, 2004. Proceedings of 2004 International Conference
  on}, vol.~7, 2004, pp. 3977--3980 vol.7.

\bibitem{kawaguchi2003iris}
T.~Kawaguchi and M.~Rizon, ``Iris detection using intensity and edge
  information,'' \emph{Pattern Recognition}, vol.~36, no.~2, pp. 549--562,
  2003.

\bibitem{zhang2008new}
Y.~Zhang, N.~Sun, Y.~Gao, and M.~Cao, ``A new eye location method based on ring
  gabor filter,'' in \emph{Automation and Logistics, 2008. ICAL 2008. IEEE
  International Conference on}.\hskip 1em plus 0.5em minus 0.4em\relax IEEE,
  2008, pp. 301--305.

\bibitem{heyma-11}
T.~Heyman, V.~Spruyt, and A.~Ledda, ``{3D Face Tracking and Gaze Estimation
  Using a Monocular Camera},'' in \emph{Proceedings of the 2nd International
  Conference on Positioning and Context-Awareness (PoCA 2011)}, Brussels,
  Belgium, 2011, p. 23{\textendash}28.

\bibitem{Choi2011}
I.~Choi, S.~Han, and D.~Kim, ``Eye detection and eye blink detection using
  adaboost learning and grouping,'' in \emph{Computer Communications and
  Networks (ICCCN), 2011 Proceedings of 20th International Conference on},
  2011, pp. 1 --4.

\bibitem{arai2011comparative}
K.~Arai and R.~Mardiyanto, ``Comparative study on blink detection and gaze
  estimation methods for hci, in particular, gabor filter utilized blink
  detection method,'' in \emph{Information Technology: New Generations (ITNG),
  2011 Eighth International Conference on}.\hskip 1em plus 0.5em minus
  0.4em\relax IEEE, 2011, pp. 441--446.

\bibitem{grauman2003communication}
K.~Grauman, M.~Betke, J.~Lombardi, J.~Gips, and G.~Bradski, ``Communication via
  eye blinks and eyebrow raises: Video-based human-computer interfaces,''
  \emph{Universal Access in the Information Society}, vol.~2, no.~4, pp.
  359--373, 2003.

\bibitem{faceTracking}
M.~Kinect, ``Kinect face tracking,''
  http://msdn.microsoft.com/en-us/library/jj130970.aspx, 2013.

\bibitem{edwards1998interpreting}
G.~J. Edwards, C.~J. Taylor, and T.~F. Cootes, ``Interpreting face images using
  active appearance models,'' in \emph{Automatic Face and Gesture Recognition,
  1998. Proceedings. Third IEEE International Conference on}.\hskip 1em plus
  0.5em minus 0.4em\relax IEEE, 1998, pp. 300--305.

\bibitem{ahlberg2001candide}
J.~Ahlberg, ``Candide-3-an updated parameterised face,'' 2001.

\bibitem{westeyn2003georgia}
T.~Westeyn, H.~Brashear, A.~Atrash, and T.~Starner, ``Georgia tech gesture
  toolkit: supporting experiments in gesture recognition,'' in
  \emph{Proceedings of the 5th international conference on Multimodal
  interfaces}.\hskip 1em plus 0.5em minus 0.4em\relax ACM, 2003, pp. 85--92.

\bibitem{lee1999hmm}
H.-K. Lee and J.-H. Kim, ``An hmm-based threshold model approach for gesture
  recognition,'' \emph{Pattern Analysis and Machine Intelligence, IEEE
  Transactions on}, vol.~21, no.~10, pp. 961--973, 1999.

\bibitem{wang2006hidden}
S.~B. Wang, A.~Quattoni, L.-P. Morency, D.~Demirdjian, and T.~Darrell, ``Hidden
  conditional random fields for gesture recognition,'' in \emph{Computer Vision
  and Pattern Recognition, 2006 IEEE Computer Society Conference on},
  vol.~2.\hskip 1em plus 0.5em minus 0.4em\relax IEEE, 2006, pp. 1521--1527.

\bibitem{murakami1991gesture}
K.~Murakami and H.~Taguchi, ``Gesture recognition using recurrent neural
  networks,'' in \emph{Proceedings of the SIGCHI conference on Human factors in
  computing systems: Reaching through technology}.\hskip 1em plus 0.5em minus
  0.4em\relax ACM, 1991, pp. 237--242.

\bibitem{young2002htk}
S.~Young, G.~Evermann, M.~Gales, T.~Hain, D.~Kershaw, X.~Liu, G.~Moore,
  J.~Odell, D.~Ollason, D.~Povey \emph{et~al.}, ``The htk book,''
  \emph{Cambridge University Engineering Department}, vol.~3, 2002.

\bibitem{CCD}
C.~car driving, ``City car driving - car driving simulator, car game,''
  http://citycardriving.com/, 2013.

\end{thebibliography}

\end{document}